\documentclass[10pt,twocolumn,letterpaper]{article}

\usepackage{cvpr}
\usepackage{times}
\usepackage{epsfig}
\usepackage{graphicx}
\usepackage{amsmath}
\usepackage{amssymb}
\usepackage{enumitem}
\usepackage{booktabs}     
\usepackage{array}        
\usepackage{makecell}     
\usepackage{pifont}
\newcommand{\cmark}{\ding{51}} 
\newcommand{\xmark}{\ding{55}} 
\usepackage{bbm}
\usepackage{placeins}
 \usepackage{nicematrix}
\usepackage[table]{xcolor}
\usepackage{multirow}
\usepackage{arydshln}
\usepackage{tikz,cite}
\usepackage[breaklinks,colorlinks]{hyperref}

\renewcommand{\arraystretch}{1.00}

\newcolumntype{C}[1]{>{\centering\arraybackslash}m{#1}}
 
\definecolor{softblue}{RGB}{232,240,254}
\definecolor{softgreen}{RGB}{233,245,236}
\definecolor{softorange}{RGB}{252,239,221}
\definecolor{softpurple}{RGB}{241,236,253}
\definecolor{softgray}{RGB}{245,246,248}
\definecolor{adairblue}{HTML}{0B6BA8}
\definecolor{arcorange}{HTML}{D55E00}
\definecolor{vitgreen}{HTML}{1B9E77}
\definecolor{swinpurple}{HTML}{8E5EA2}
\definecolor{metricgray}{RGB}{255,255,255}
\definecolor{scoutred}{HTML}{C1121F}
\definecolor{styleclipcyan}{HTML}{00B4D8}
\definecolor{wplusgold}{HTML}{FFB703}
\newcommand{\pmark}{\raisebox{0.5pt}{\scalebox{0.8}{$\sim$}}}

\newcommand{\imgcellfile}[2][1.64cm]{%
  \includegraphics[width=#1,height=#1]{#2}%
}
 
\newlength{\smallrowht}
\newlength{\deltarowht}
\newlength{\cossimrowht}
\newcommand{\inputimagelabel}{%
  \parbox[c][\smallrowht][c]{2.00cm}{%
    \centering\scriptsize Input image%
  }%
}

\newcommand{\targetupper}[1]{%
  \parbox[c][1.64cm][c]{2.00cm}{\centering
    \vspace{-1.5mm}%
    \includegraphics[width=1.64cm,height=1.64cm,keepaspectratio]{#1}%
  }%
}

\newcommand{\targetlower}[4]{%
  \parbox[c][\dimexpr 1.64cm + 2\deltarowht\relax][c]{2.00cm}{\centering
    \vspace{7pt}
    {\scriptsize \(c=\) \,\emph{#1}\par}
    \vspace{5pt}
    {\scriptsize \(n=\) \,#2\par}
    \vspace{5pt}
    {\scriptsize \(S_n(c)=\) \,#3\par}
    \vspace{5pt}
    {\scriptsize \(A_n(c)=\) \,#4\par}
  }%
}

\newcommand{\targetcossim}{%
  \parbox[c][\cossimrowht][c]{2.00cm}{\centering\scriptsize Cosine similarity}%
}
\newcommand{\alphacell}[1]{%
  \parbox[c][\smallrowht][c]{\linewidth}{\centering\scriptsize $#1$}%
}
\newcommand{\deltacell}[1]{%
  \parbox[c][\deltarowht][c]{\linewidth}{\centering\scriptsize $\Delta_c$:~#1}%
}
\newcommand{\cossimcell}[1]{%
  \parbox[c][\cossimrowht][c]{\linewidth}{\centering\scriptsize $\theta_c$:~#1}%
}

\newlength{\rowskip}
\newlength{\baldsmallrowht}
\newlength{\baldmetricrowht}
\begin{document}

\title{\vspace{-3mm}SCOUT: Semantic Concept Discovery for Open-Vocabulary Editing of Face Recognition Templates}

\author{Leon Todorov$^{1}$ \quad Peter Rot$^{1}$ \quad Peter Peer$^{2}$ \quad
Vitomir \v{S}truc$^{1}$ \quad Klemen Grm$^{1}$\\
{\small University of Ljubljana: $^{1}$Faculty of Electrical Engineering \quad $^{2}$Faculty of Computer and Information Science}\\
{\tt\small Leon.Todorov@fe.uni-lj.si \quad Peter.Rot@fe.uni-lj.si \quad Peter.Peer@fri.uni-lj.si}\\
{\tt\small Vitomir.Struc@fe.uni-lj.si \quad Klemen.Grm@fe.uni-lj.si}
}

\maketitle
\thispagestyle{empty}

\begin{abstract}
Face recognition templates are compact identity representations, yet they also encode rich semantic information about facial appearance. Prior work has shown that templates can be inverted to images or indirectly manipulated through image-editing pipelines, but direct semantic editing in template space remains largely unexplored. Existing interpretability methods for face recognition often rely on manual neuron inspection or predefined attribute labels, limiting scalability and semantic flexibility. To address this gap, we propose \textbf{SCOUT} (\textbf{S}emantic \textbf{C}oncept Discovery for \textbf{O}pen-Vocab\textbf{U}lary Editing of Face Recognition \textbf{T}emplates), an end-to-end framework for discovering and directly manipulating semantic concepts in face recognition templates using mechanistic interpretability. SCOUT learns sparse template representations, generates semantic hypotheses for latent features from natural-language descriptions, and validates their stability. The resulting features act as controllable semantic directions for direct editing, avoiding costly edit--re-encode pipelines. Experiments with face recognition models using CNN, ViT, and Swin backbones show that SCOUT discovers interpretable concepts beyond standard attribute labels and enables controllable, identity-aware template manipulation with negligible impact on identity matching. We further show that edited templates can subsequently be decoded with independent inversion models for visualization and evaluation.
\end{abstract}

\section{Introduction}

\begin{figure}[!t]
    \centering
    \includegraphics[width=0.84\columnwidth]{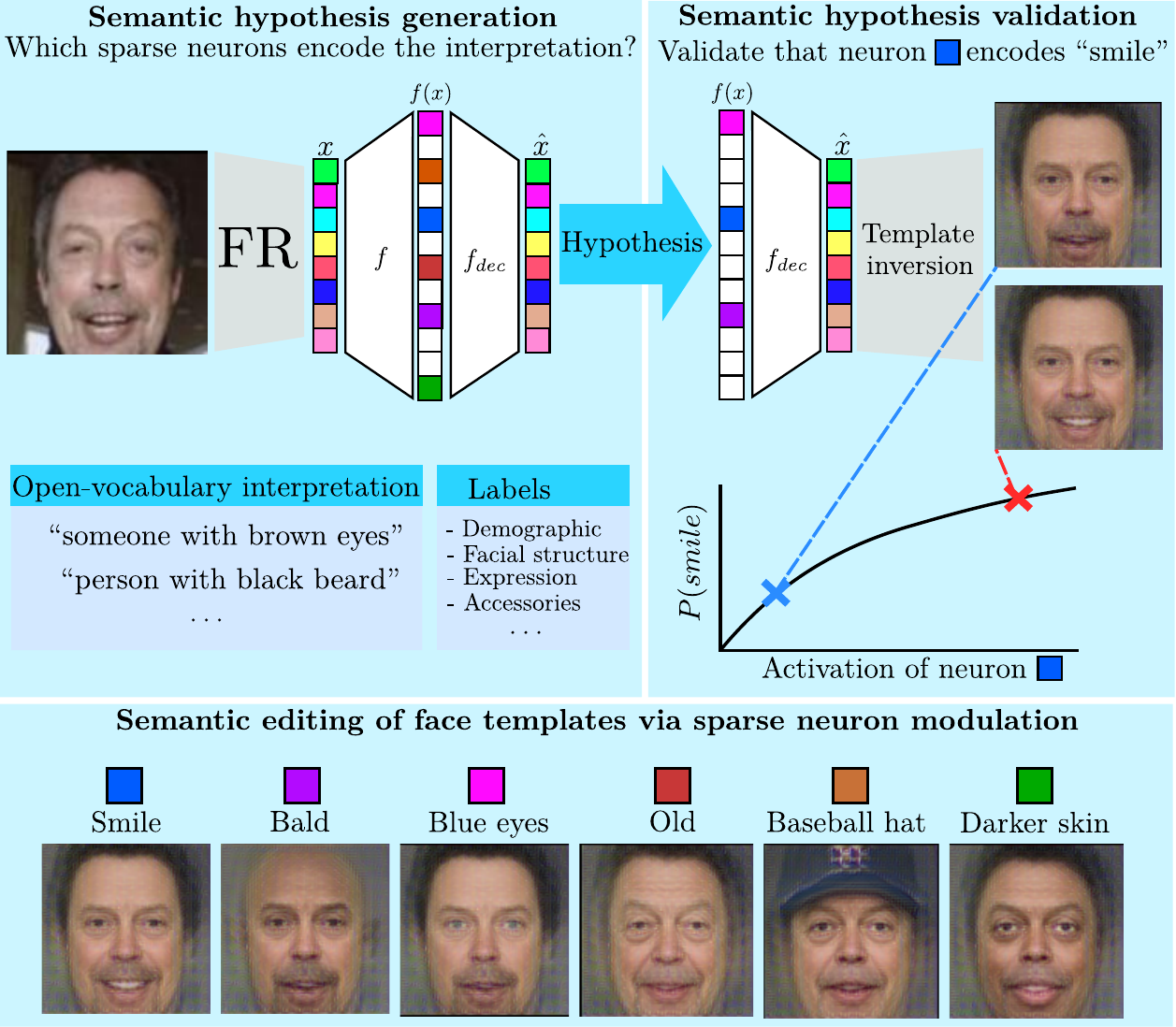}
    \caption{SCOUT automatically discovers and validates semantic interpretations of sparse neurons in the sparse autoencoder (SAE) using face attributes defined by free-form text prompts or predefined labels. Once stable neuron–attribute associations are identified through inversion-based verification, the corresponding sparse features are used to directly manipulate semantic attributes in template space for efficient template augmentation.}
    \label{fig:highlevel_banner}
    \vspace{-3mm}
\end{figure}

Face recognition (FR) systems represent identity through compact vector representations (templates) that enable efficient matching and storage~\cite{kim2022adaface}.~Although templates are trained for discriminative identity inference, a growing body of work shows they encode substantially richer facial information than identity alone~\cite{terhorst2020beyond}. Modern template inversion and embedding-conditioned synthesis can also reconstruct photorealistic faces from embeddings with high identity fidelity~\cite{shahreza2024vulnerability,papantoniou2024arc2face}. Beyond inversion, several studies have explored the semantic organization of FR embedding space. Direction-discovery methods~\cite{plesh2024discovering} identify interpretable axes corresponding to facial regions (e.g., eyes, nose, lower face) or nuisance factors (pose, illumination, expression), often visualizing linear traversals by decoding embeddings back into pixel space. In parallel, language-based approaches such as CLIP-SMU~\cite{manojlovska2025interpreting} aim to translate template content into natural-language descriptions by aligning templates with CLIP-derived text representations over predefined attribute descriptions. These works establish that FR templates reside in an embedding space that exhibits meaningful semantic structure and can be probed through either generative inversion or semantic alignment.

Due to this semantic structure, the information content in FR templates can easily be manipulated, and certain characteristics independent of identity can be suppressed, strengthened, altered, or even added directly in the embedding space. This manipulation process, referred to as \textit{template editing}, is critical in many scenarios, including privacy-preserving techniques that suppress sensitive attributes, e.g., sex, age, or ethnicity~\cite{melzi2023multi,rot2024aspecd}, robustness and bias analysis through controlled perturbations of embeddings~\cite{serna2026discoveringintersectionalbiasdirectional,leroy2025attributes}, synthetic identity generation and augmentation from manipulated embeddings~\cite{shahreza2024hyperface}, and improved understanding of the encoding process of FR models.   

However, existing methods for semantic manipulation of templates remain largely indirect or task-specific. Most fine-grained changes are still implemented by editing in image space or a generative latent space and then re-encoding~\cite{bai2024real,jiang2021talk} the edited image into the embedding space of the corresponding FR model. While feasible and effective, such processes introduce computational overhead and couple the manipulation operation to the failure modes of the synthesis pipelines used. Template-only transformations, such as those used to suppress attributes (e.g., age or gender) in privacy-preservation methods, are typically limited to coarse facial attributes~\cite{dai2026clip,grimmer2024ladimo,rot2024aspecd}, while general mechanisms for fine-grained, open-vocabulary semantic editing directly in standard FR embeddings, without image-level inversion, remain largely missing from the literature. Here, \emph{open-vocabulary} refers to the ability to define semantic edits through free-form natural-language descriptions rather than a closed set of predefined (and often binary) attributes, enabling manipulation of concepts such as \textit{``wearing a baseball hat''} or \textit{``having blue eyes''}, as illustrated in Figure~\ref{fig:highlevel_banner}.

Addressing this gap requires not only a template-space editing operator, but also \textit{an automatic way} to identify which components of an embedding correspond to meaningful, controllable semantics.~Here, recent progress in \emph{mechanistic interpretability} provides a compelling direction. Mechanistic interpretability aims to explain a model in terms of internal computational components and circuits, identifying which features or mechanisms inside a representation drive particular behaviors or outputs~\cite{cunningham2023sparse}. Sparse autoencoders (SAEs) have emerged as an unsupervised tool for decomposing dense neural representations into sparse, more monosemantic features~\cite{gao2024scaling}. In this work, we build on these advances and introduce a template-only system for semantic manipulation of face-recognition embeddings under open-vocabulary control. We propose \textbf{SCOUT} (\textbf{S}emantic \textbf{C}oncept Discovery for \textbf{O}pen-Vocab\textbf{u}lary Editing of Face Recognition \textbf{T}emplates), an end-to-end framework with three main contributions:\vspace{-1mm}
\begin{itemize}[itemsep=0pt]
    \item An automatic method for discovering interpretable semantic features in learned sparse representations of face recognition templates, guided by natural-language descriptions (e.g., using CLIP).\vspace{-1mm}
    \item A direct template-space editing mechanism that uses the discovered features as controls for fine-grained semantic manipulation, without image-level editing or re-encoding of the input face.\vspace{-1mm}
    \item A method for creating controlled variations of face templates by modifying semantic attributes (e.g., adding a beard or changing the hairstyle), which can subsequently be visualized through template inversion methods such as Arc2Face.
\end{itemize}\vspace{-2mm}


\section{Related work}
\label{sec:related_work}
\subsection{Discovery of Interpretable Concepts}
\vspace{-1mm}
Early work on interpretable structure in learned embeddings relied on post-hoc methods such as Network Dissection~\cite{Bau2017NetworkDissection} and TCAV~\cite{kim2018tcav}, which typically require predefined concepts and human supervision. More recent approaches aim to automate concept discovery. MAIA~\cite{shaham2024multimodal} iteratively generates and tests hypotheses about neuron behavior using vision-language models and interpretability tools, while Prisma~\cite{joseph2025prisma} supports large-scale mechanistic interpretability through sparse autoencoders (SAEs). In parallel, SAE-based methods such as MSAE~\cite{zaigrajew2025interpreting} and Revelio~\cite{Kim_2025_ICCV}, as well as open-vocabulary sparse decompositions such as SpLiCE~\cite{bhalla2024interpreting}, show that semantically meaningful structure can be recovered in general vision representations. However, these methods target broad vision settings. Face recognition is more structured: faces are typically aligned and normalized, and recent work shows~\cite{plesh2024discovering,leroy2025attributes} that FR embeddings exhibit geometric organization with respect to interpretable facial and image attributes. This suggests that semantic factors in FR templates may be more stable and analyzable than in generic visual representations, although existing vision-oriented methods are not adapted to the structured geometry of face-recognition templates.

Only limited work has explored mechanistic interpretability in biometrics. Prior studies have examined what information is encoded in face templates~\cite{terhorst2020beyond}, their geometric relation to attributes~\cite{leroy2025attributes}, and, more recently, unsupervised discovery of semantically coherent directions for bias analysis~\cite{serna2026discoveringintersectionalbiasdirectional}.~FaceMINT~\cite{rot2025facemint}, the work most closely related to ours, enables sparse feature analysis and feature-level interventions for face-recognition representations. However, semantic interpretation still relies largely on manual inspection, for example, through top-activating samples or human validation. Overall, biometrics still lacks end-to-end pipelines for open-vocabulary semantic discovery, sparse decomposition, validation, and direct manipulation of biometric templates. A summary of existing approaches and their characteristics is presented in Table~\ref{tab:method-comparison}.

\begin{table}[tb]
\caption{Overview of methods for semantic concept discovery in face recognition templates. \cmark: supported, \xmark: not supported, \pmark: partially supported or not the primary focus.\vspace{1mm}} 
\label{tab:method-comparison}
\centering
\scriptsize
\setlength{\tabcolsep}{2pt}
\renewcommand{\arraystretch}{1.05}
\resizebox{\linewidth}{!}{%
\begin{NiceTabular}{@{}>{\centering\arraybackslash}m{0.42cm}
>{\raggedright\arraybackslash}m{2.35cm} c c c c c@{}}[color-inside]
\toprule
\addlinespace[1pt]
&
\textbf{Works} &
\shortstack{\textbf{Open-}\\\textbf{vocabulary}} &
\shortstack{\textbf{Sparse}\\\textbf{decomposition}} &
\shortstack{\textbf{Automatic}\\\textbf{discovery}} &
\shortstack{\textbf{Validation /}\\\textbf{intervention}} &
\shortstack{\textbf{Direct}\\\textbf{editing}} \\
\addlinespace[2pt]
\midrule
\addlinespace[1pt]
\Block[tikz={fill=blue!12,draw=none}]{7-1}
{\rotatebox[origin=c]{90}{\scriptsize\textit{General vision}}}
& Net Dissection~\cite{Bau2017NetworkDissection} & \xmark & \xmark & \xmark & \pmark & \xmark \\
& TCAV~\cite{kim2018tcav} & \xmark & \xmark & \xmark & \pmark & \xmark \\
& MAIA~\cite{shaham2024multimodal} & \cmark & \xmark & \cmark & \cmark & \xmark \\
& Prisma~\cite{joseph2025prisma} & \xmark & \cmark & \pmark & \pmark & \xmark \\
& MSAE~\cite{zaigrajew2025interpreting} & \xmark & \cmark & \pmark & \pmark & \xmark \\
& Revelio~\cite{Kim_2025_ICCV} & \xmark & \cmark & \pmark & \pmark & \xmark \\
& SpLiCE~\cite{bhalla2024interpreting} & \cmark & \cmark & \pmark & \pmark & \cmark \\ \addlinespace[1pt]
\midrule
\addlinespace[1pt]
\Block[tikz={fill=green!12,draw=none}]{5-1}
{\rotatebox[origin=c]{90}{\scriptsize\textit{Biometrics}}}
& Terhorst \etal~\cite{terhorst2020beyond} & \xmark & \xmark & \cmark & \xmark & \xmark \\
& Leroy \etal~\cite{leroy2025attributes} & \xmark & \xmark & \cmark & \xmark & \xmark \\
& CLIP-SMU~\cite{manojlovska2025interpreting} & \cmark & \xmark & \cmark & \xmark & \xmark \\
& DeepFace Decoder~\cite{plesh2024discovering} & \xmark & \pmark & \xmark & \cmark & \xmark \\
& FaceMINT~\cite{rot2025facemint} & \xmark & \cmark & \xmark & \cmark & \pmark \\\addlinespace[1pt]
\midrule
\addlinespace[1pt]
\rowcolor{gray!12}
& \textbf{SCOUT (ours)} & \cmark & \cmark & \cmark & \cmark & \cmark \\
\bottomrule
\end{NiceTabular}
}
\vspace{-4.55mm}
\end{table}

\subsection{Semantic Manipulation of Face Templates}

Existing approaches for the semantic manipulation of face templates can be grouped into indirect \emph{edit--re-encode} pipelines, inversion-based manipulation, and direct template-space transformations such as privacy-preserving suppression, morphing, or semantic direction control. 

\vspace{0.5mm}\noindent\textbf{Image~Editing~and~Re-Encoding.}~Techniques from this group perform semantic manipulation outside the FR template space, either by directly editing the face image or by modifying a latent representation of the image in a generative model. Image-level methods apply semantic transformations to the input face~\cite{bai2024real,jiang2021talk} and then pass the edited result through an FR model to obtain a modified template. Latent-space methods, in contrast, first project the face image into a generative latent space~\cite{richardson2021encoding}, apply semantic edits in that space~\cite{shen2020interfacegan}, and then decode and re-encode the edited image. Together, these approaches form a standard \emph{edit--re-encode} pipeline. In many cases, identity preservation is explicitly enforced using pretrained FR models such as ArcFace~\cite{bai2024real,patashnik2021styleclip}. These methods offer rich semantic control, but the manipulation remains indirect: edits are applied outside the template space and only later reflected in the biometric representation. As a result, they incur the cost of inversion, editing, generation, and feature extraction, while strong edits may also introduce identity drift.


\vspace{0.5mm}\noindent\textbf{Template Inversion with Guided Decoding.}~Another line of work uses FR templates as inputs to generative models that reconstruct face images.~Methods such as Arc2Face~\cite{papantoniou2024arc2face} and Face Adapter~\cite{shahreza2025faceadapter} show that high-quality face images can be reconstructed from templates, while CLIP-FTI~\cite{dai2026clip} adds an auxiliary semantic signal, such as a text prompt, to guide the reconstruction toward a desired attribute. In this case, the template itself is not edited directly. Instead, the semantic content is manipulated indirectly by decoding the same template in a specific way, using the additional guidance signal. Related work also shows that such reconstructions can be used to inspect and visualize semantic directions in template space~\cite{plesh2024discovering}. These methods show that FR templates contain rich semantic information that can be exposed through guided reconstruction. However, the semantic change is still produced by the decoder, rather than by an explicit and controllable transformation of the template itself.


\vspace{0.5mm}\noindent\textbf{Direct~Template~Transformation.}~Several approaches modify FR templates directly while preserving their compatibility with the original FR model. Such transformations are commonly used to change specific properties of the template, for example by suppressing sensitive or soft-biometric information. PFRNet~\cite{bortolato2020learning}, for example, learns disentangled representations that remove selected attributes while preserving identity, and follow-up methods such as Multi-IVE~\cite{melzi2023multi} and ASPECD~\cite{rot2024aspecd} extend this idea to multiple or categorical attributes. These methods show that FR templates can be transformed without image reconstruction or changes to the FR model. However, the transformations are typically designed for a predefined set of attributes, rather than flexible, open-vocabulary semantic editing.

\vspace{0.5mm}\noindent\textbf{Synthetic Identity Generation.}
Recent work also manipulates face embeddings to synthesize new identities: HyperFace~\cite{shahreza2024hyperface} formulates dataset creation as a packing problem on the face-embedding hypersphere, while diffusion-based ID-Booth~\cite{tomasevic2025idbooth} or ID-Sync \cite{JernejFG2026} target identity-consistent generation for data augmentation. These methods, however, focus on large-scale synthetic face generation rather than flexible semantic control over existing templates.


\begin{figure*}[t]
    \centering
    \includegraphics[width=0.94\textwidth]{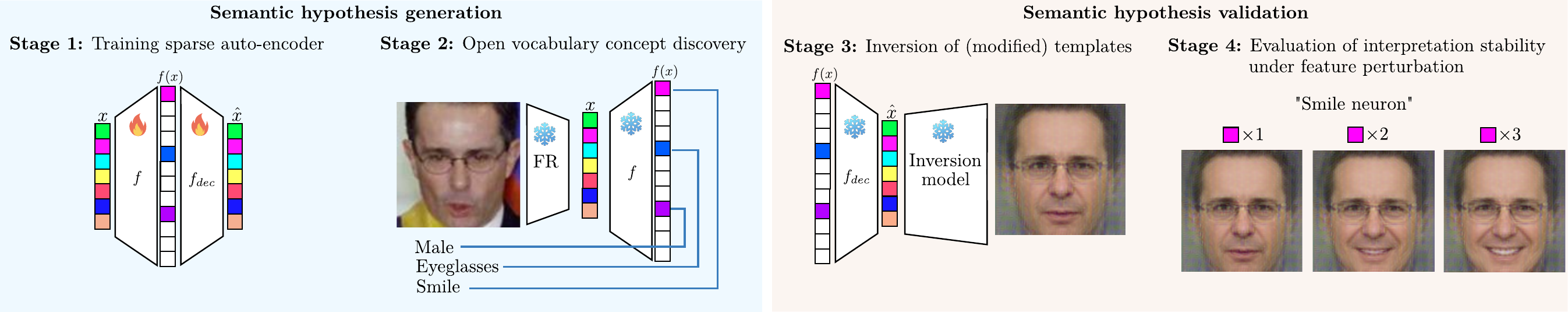}
    \caption{High-level overview of SCOUT, the proposed framework for discovering and manipulating semantic concepts directly in face recognition template space. The framework consists of four stages: (1) training a sparse autoencoder on face templates, (2) assigning latent neurons to semantic concepts using labeled or open-vocabulary attributes, (3) reconstructing faces via template inversion, and (4) validating the discovered concepts through neuron-level interventions, where individual sparse features are perturbed to measure their semantic effect.}
    \label{fig:highlevel}
    \vspace{-3mm}
\end{figure*}

\section{Methodology}
\label{sec:methodology}

In this section, we present SCOUT, the proposed framework for discovering and manipulating semantic concepts directly in face recognition template space. The key idea behind SCOUT is to first decompose dense FR templates into sparse, interpretable features and then identify which of these features correspond to user-defined semantic concepts. Once such concept-linked features are found, they can be used as direct controls for modifying the original template representation, while inversion models provide a way to visually inspect the resulting semantic changes in image space. As illustrated in Figure~\ref{fig:highlevel}, SCOUT proceeds in four stages: sparse decomposition of FR templates with an SAE (Stage~1), concept discovery producing candidate concept--feature associations (Stage~2), inversion-based visualization of these associations in image space (Stage~3), and validation through feature-level interventions that quantify both semantic change and identity preservation (Stage~4).


\subsection{Sparse Autoencoder Training (Stage 1)}\label{subsec:sae-training}
Given an aligned and cropped face image $I$, an FR model produces a template $\mathbf{x}
\in \mathbb{R}^{d}$. Using a BatchTopK SAE~\cite{bussmann2024batchtopk}, we map $
\mathbf{x}$ to an overcomplete latent representation $\mathbf{z} = f(\mathbf{x}) \in
\mathbb{R}^{d_{\mathrm{SAE}}}$, where $d_{\mathrm{SAE}} \gg d$ sets the dictionary size.
During training, BatchTopK sparsification imposes a global activation budget $k$ over each
minibatch, retaining only the strongest latent activations and thereby controlling the
effective sparsity of $\mathbf{z}$. The input is reconstructed as $\hat{\mathbf{x}} =
f_{\mathrm{dec}}(\mathbf{z})$, and the SAE is trained to minimize the reconstruction error
between $\hat{\mathbf{x}}$ and $\mathbf{x}$. At inference time, the batchwise selection is
replaced by a learned global activation threshold $\tau$, enabling sample-wise encoding
without batch context.~Once trained, the SAE represents each dense FR template as a sparse set of feature
activations. Each feature is paired with a learned vector in the original template
space, referred to as \emph{dictionary vector}, which specifies how that feature contributes to
reconstructing the input template. These template-space dictionary vectors provide the
basic units for concept discovery and template-level intervention in the following stages.


\subsection{Concept Discovery (Stage 2)}
\label{subsec:concept-definition}

The goal of this stage is to assign semantic interpretations to the sparse features in
$\mathbf{z}$.~For a queried concept $c$, we construct positive and negative exemplar
sets, rank sparse features according to their selectivity for the concept, and finally
verify the resulting concept-feature associations using top-activating samples.~The output is a set of semantic hypotheses carried forward for downstream validation.

\vspace{0.5mm}\noindent\textbf{Concept Exemplar Set Construction.}~For a concept $c$, let $\mathcal{D}^{+}_{c}$ and $\mathcal{D}^{-}_{c}$ denote images
that exhibit and lack the concept.~These sets can be obtained in two
ways: (1) in the \emph{closed-vocabulary} setting via a supervised attribute
classifier over a fixed taxonomy, and (2) in the \emph{open-vocabulary} setting via a
vision-language model queried with natural-language prompts. For open-vocabulary concepts, we query the vision-language model with two disjoint prompt banks: one used for Stage~2 discovery and feature validation, and one reserved for measuring interventions in Stage~4. Thus, prompts used to select a direction are not reused to evaluate it. In both cases, each training
image $I$ receives a concept confidence score used to select the top- and
bottom-ranked images as $\mathcal{D}^{+}_{c}$ and $\mathcal{D}^{-}_{c}$.

\vspace{0.5mm}\noindent\textbf{Selectivity-Based Feature Ranking.}~Once the exemplar sets are defined, we evaluate which features in $\mathbf{z}$ respond
selectively to concept $c$. Let $z_n(I)$ denote the $n$-th component of $\mathbf{z}$ for
image $I$. We compute the empirical activation rates
\begin{equation}
r_n^{+}(c)=\frac{1}{|\mathcal{D}^{+}_{c}|}\sum_{I\in\mathcal{D}^{+}_{c}}\mathbbm{1}
[z_n(I)>\tau],
\end{equation}
\begin{equation}
r_n^{-}(c)=\frac{1}{|\mathcal{D}^{-}_{c}|}\sum_{I\in\mathcal{D}^{-}_{c}}\mathbbm{1}
[z_n(I)>\tau],
\end{equation}
and define the selectivity score as
\begin{equation}
S_n(c)=r_n^{+}(c)-r_n^{-}(c)\in[-1,1].
\end{equation}
Values near zero indicate weak discrimination between the two sets. High values indicate
that feature $n$ is active predominantly on concept-positive samples, whereas negative
values indicate an association with concept absence. Features are ranked by $S_n(c)$, and
the top candidates are retained for subsequent validation and intervention.~Using
activation rates rather than activation magnitudes captures whether a feature is reliably
present for a concept, consistent with the sparse nature of the representation.

\vspace{0.5mm}\noindent\textbf{Concept Agreement.}~The selectivity score measures differential activation, but does not verify whether the
assigned concept label accurately describes what a feature encodes. We therefore validate
each candidate association by checking whether the images that most strongly activate
feature $n$ are also classified as concept-positive.

For each candidate feature $n$, let $\mathcal{T}_{n}$ denote the set of highest-activating
images collected over the full training corpus. We define the concept agreement as
\begin{equation}
A_n(c)=\frac{1}{|\mathcal{T}_{n}|}\sum_{I\in\mathcal{T}_{n}}\mathbbm{1}\!
\left[s_c^{\mathrm{disc}(I)}>\tfrac{1}{2}\right],
\end{equation}
where $s_c^{\mathrm{disc}}(I)$ is the discovery confidence score used to form the exemplar sets. In the open-vocabulary setting, it is computed with the discovery prompt bank. High values
of $A_n(c)$ indicate that the images most strongly driving feature $n$ are consistently
assigned to concept $c$, whereas mismatched high-activating samples act as distractors. A
minimum concept-agreement threshold is then used to discard weakly aligned features and
obtain the final concept--feature associations.

\subsection{Concept Visualization via Inversion (Stage 3)}

The concept--feature associations identified in Stage~2 define semantic hypotheses about
individual sparse features. Stage~3 uses inversion to inspect these hypotheses in image
space. Given an FR template $\mathbf{x}$,
the DeepFace Decoder (DFD)~\cite{krizaj2024deepfacedecoder,plesh2024discovering} provides a lightweight
regression-based reconstruction, while Arc2Face~\cite{papantoniou2024arc2face} provides an
independently trained diffusion-based view in the ARC-iR100 space. These decoders are
auxiliary, i.e., they visualize original or intervened templates but neither produce the
template-space intervention nor feed a reconstructed image back through the FR encoder.
Agreement across the two therefore reduces the likelihood that an observed semantic effect
is decoder-specific.

\begin{figure*}[!t]
\centering
\refstepcounter{figure}\label{fig:qualitative_backbones}
{%
\setlength{\tabcolsep}{0pt}%
\renewcommand{\arraystretch}{1.0}%
\newlength{\rbImgW}\setlength{\rbImgW}{1.64cm}%
\newlength{\rbLabelW}\setlength{\rbLabelW}{0.46cm}%
\newlength{\rbMetricH}\setlength{\rbMetricH}{0.28cm}%
\newlength{\rbAlphaH}\setlength{\rbAlphaH}{0.24cm}%
\newlength{\rbRightW}\setlength{\rbRightW}{6.20cm}%
\newcommand{\rbAlpha}[1]{\parbox[c][\rbAlphaH][c]{\rbImgW}{\centering\scriptsize $\alpha = #1$}}%
\newcommand{\rbImg}[1]{\includegraphics[width=\rbImgW,height=\rbImgW]{#1}}%
\newcommand{\rbDelta}[1]{\parbox[c][\rbMetricH][c]{\rbImgW}{\centering\scriptsize $\Delta_c$: #1}}%
\newcommand{\rbTheta}[1]{\parbox[c][\rbMetricH][c]{\rbImgW}{\centering\scriptsize $\theta_c$: #1}}%
\newcommand{\rbRotLabel}[2]{%
  \parbox[c][\dimexpr\rbAlphaH+\rbImgW+2\rbMetricH\relax][c]{\rbLabelW}{%
    \centering\rotatebox[origin=c]{90}{\scriptsize\bfseries\textcolor{#1}{#2}}%
  }%
}%

\newsavebox{\rbLeftBox}
\savebox{\rbLeftBox}{%
\begin{tabular}{@{}m{5.18cm}@{\hspace{0.34cm}}m{5.18cm}@{}}
\begin{tabular}{@{}m{\rbLabelW}@{}m{\rbImgW}m{\rbImgW}m{\rbImgW}@{}}
\multirow{4}{*}[0.1cm]{\rbRotLabel{adairblue}{ADA-iR50}}
& \rbAlpha{0.00} & \rbAlpha{0.20} & \rbAlpha{0.40}\\[-0.05em]
& \rbImg{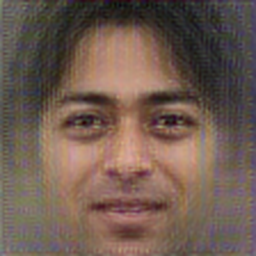}
& \rbImg{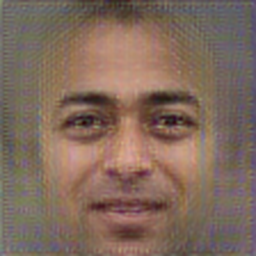}
& \rbImg{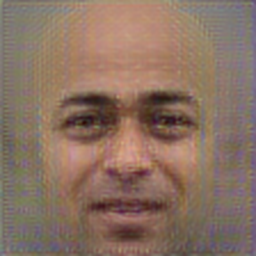}\\[-0.02em]
& \rbDelta{0.000} & \rbDelta{0.274} & \rbDelta{0.585}\\[-0.10em]
& \rbTheta{1.000} & \rbTheta{0.980} & \rbTheta{0.925}
\end{tabular}
&
\begin{tabular}{@{}m{\rbLabelW}@{}m{\rbImgW}m{\rbImgW}m{\rbImgW}@{}}
\multirow{4}{*}[0.1cm]{\rbRotLabel{arcorange}{ARC-iR100}}
& \rbAlpha{0.00} & \rbAlpha{0.20} & \rbAlpha{0.40}\\[-0.05em]
& \rbImg{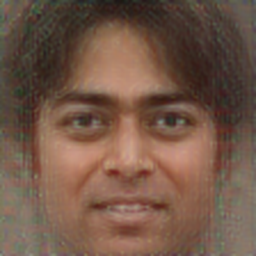}
& \rbImg{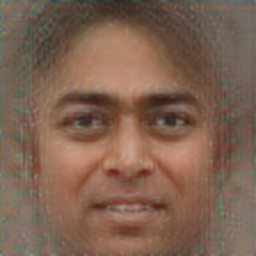}
& \rbImg{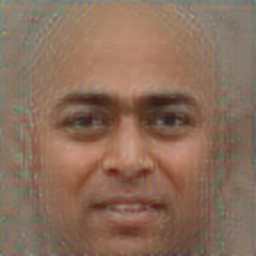}\\[-0.02em]
& \rbDelta{0.000} & \rbDelta{0.290} & \rbDelta{0.576}\\[-0.10em]
& \rbTheta{1.000} & \rbTheta{0.980} & \rbTheta{0.928}
\end{tabular}\\[0.08cm]
\begin{tabular}{@{}m{\rbLabelW}@{}m{\rbImgW}m{\rbImgW}m{\rbImgW}@{}}
\multirow{4}{*}[1.2cm]{\rbRotLabel{vitgreen}{ADA-ViT}}
& \rbImg{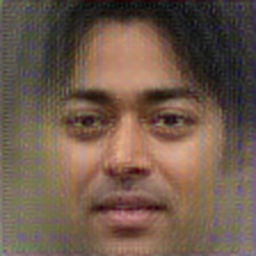}
& \rbImg{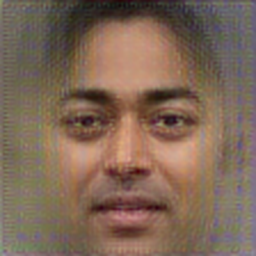}
& \rbImg{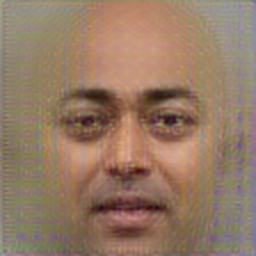}\\[-0.02em]
& \rbDelta{0.000} & \rbDelta{0.220} & \rbDelta{0.754}\\[-0.10em]
& \rbTheta{1.000} & \rbTheta{0.981} & \rbTheta{0.929}
\end{tabular}
&
\begin{tabular}{@{}m{\rbLabelW}@{}m{\rbImgW}m{\rbImgW}m{\rbImgW}@{}}
\multirow{4}{*}[1.2cm]{\rbRotLabel{swinpurple}{SWIN-T}}
& \rbImg{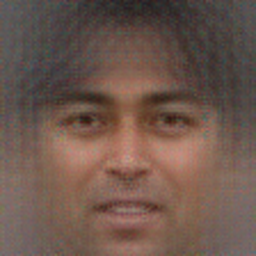}
& \rbImg{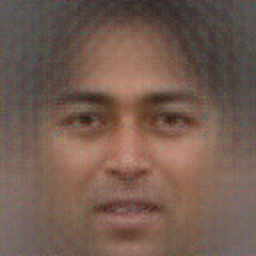}
& \rbImg{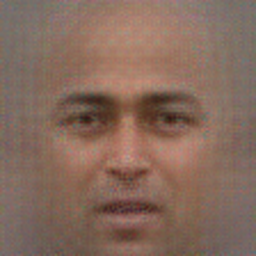}\\[-0.02em]
& \rbDelta{0.000} & \rbDelta{0.024} & \rbDelta{0.734}\\[-0.10em]
& \rbTheta{1.000} & \rbTheta{0.981} & \rbTheta{0.930}
\end{tabular}
\end{tabular}%
}
\newlength{\rbLeftH}\setlength{\rbLeftH}{\dimexpr\ht\rbLeftBox+\dp\rbLeftBox\relax}

\newsavebox{\rbTableBox}
\savebox{\rbTableBox}{%
{\scriptsize
\renewcommand{\arraystretch}{1.02}
\setlength{\tabcolsep}{0pt}
\begin{tabular}{@{}>{\raggedright\arraybackslash}p{1.40cm}>{\centering\arraybackslash}p{0.56cm}>{\centering\arraybackslash}p{0.78cm}>{\centering\arraybackslash}p{0.94cm}>{\centering\arraybackslash}p{1.12cm}>{\centering\arraybackslash}p{1.12cm}@{}}
\toprule
\textbf{Model} & \textbf{$n$} & \textbf{$S_n(c)$} & \textbf{$A_n(c)$} & \textbf{$\overline{\Delta}_c(0.2)$} & \textbf{$\overline{\Delta}_c(0.4)$}\\
\midrule
\rowcolor{softblue}
\textcolor{adairblue}{\textbf{ADA-iR50}} & $5971$ & $0.719$ & $99.9\%$ & $0.200$ & $0.527$\\
\rowcolor{softorange}
\textcolor{arcorange}{\textbf{ARC-iR100}} & $915$ & $0.572$ & $99.9\%$ & $0.197$ & $0.495$\\
\rowcolor{softgreen}
\textcolor{vitgreen}{\textbf{ADA-ViT}} & $4430$ & $0.794$ & $100.0\%$ & $0.107$ & $0.487$\\
\rowcolor{softpurple}
\textcolor{swinpurple}{\textbf{SWIN-T}} & $5713$ & $0.593$ & $92.3\%$ & $0.042$ & $0.283$\\
\bottomrule
\end{tabular}}
}
\newlength{\rbTableH}\setlength{\rbTableH}{\dimexpr\ht\rbTableBox+\dp\rbTableBox\relax}
\newlength{\rbPlotH}\setlength{\rbPlotH}{\dimexpr\rbLeftH-\rbTableH\relax}

\noindent\makebox[0.98\textwidth][c]{%
\begin{minipage}[t]{10.82cm}
\vspace*{0pt}%
\usebox{\rbLeftBox}
\end{minipage}%
\hspace{0.42cm}%
\begin{minipage}[t][\rbLeftH][t]{\rbRightW}
\vspace*{0.15cm}%
\hspace*{0.25cm}\makebox[\dimexpr\rbRightW-0.10cm\relax][r]{\usebox{\rbTableBox}}\par
\vspace{1mm}
\hspace*{-0.3cm}\includegraphics[width=1.05\linewidth]{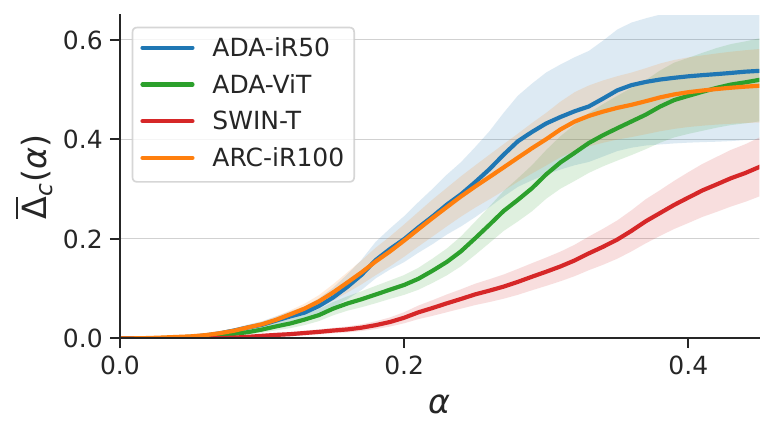}
\end{minipage}%
}%


\noindent\makebox[0.98\textwidth][c]{%
\begin{tabular}{@{}m{1.82cm}@{\hspace{0.24cm}}m{14.96cm}@{}}
\parbox[t]{1.82cm}{\centering{\scriptsize Input image\par}\vspace{-0.1em}%
\includegraphics[width=1.64cm,height=1.64cm]{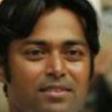}}
&
\parbox[t]{14.96cm}{\vspace*{0.07cm}\small Figure \thefigure. Full pipeline demonstration for \(c=\) \emph{bald}
across four FR backbones. Left: the same input template is manipulated at
\(\alpha\in\{0.00,0.20,0.40\}\) and decoded with the corresponding DFD;
\(\Delta_c\) is averaged across disjoint evaluation-prompt subsets, while
\(\theta_c\) denotes template cosine similarity. Right: the table reports
feature index \(n\), Stage-2 selectivity \(S_n(c)\), concept agreement
\(A_n(c)\), and evaluation-prompt mean gains at \(\alpha=0.2,0.4\).
Curves show the mean over 100 LFW templates and all evaluation-prompt subsets; shading
denotes the pointwise \(\pm1\) standard deviation across evaluation-prompt subsets.}
\end{tabular}%
}\vspace{1mm}%
}%
\end{figure*}
\begin{figure*}[!t]
\centering

\begin{subfigure}[t]{0.245\textwidth}
\centering
\includegraphics[width=\linewidth]{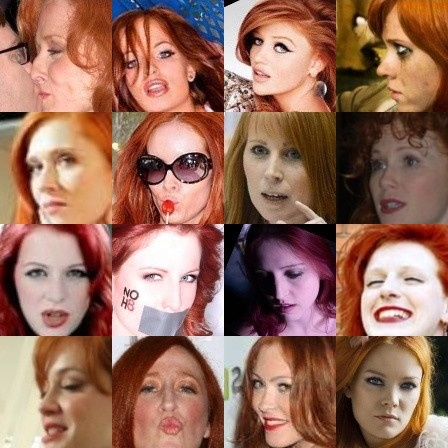}
\captionsetup{justification=centering}
\caption{\textcolor{swinpurple}{\textbf{SWIN-T}}: \(c=\) \emph{long red hair}\\\(n=6823\)\\\mbox{\(S_n(c)=0.593,\;A_n(c)=\mathrm{0.841}\).}}
\end{subfigure}\hfill
\begin{subfigure}[t]{0.245\textwidth}
\centering
\includegraphics[width=\linewidth]{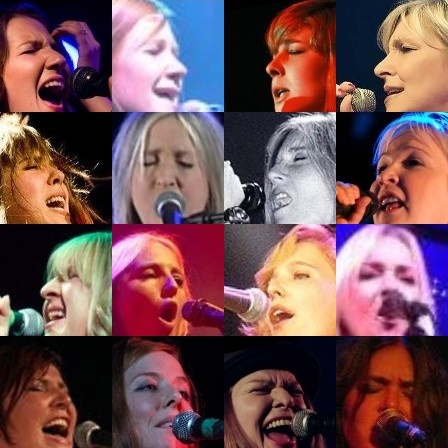}
\captionsetup{justification=centering}
\caption{\textcolor{vitgreen}{\textbf{ADA-ViT}}: \(c=\) \emph{singing}\\\(n=623\)\\\mbox{\(S_n(c)=0.514,\;A_n(c)=\mathrm{0.899}\).}}
\end{subfigure}\hfill
\begin{subfigure}[t]{0.245\textwidth}
\centering
\includegraphics[width=\linewidth]{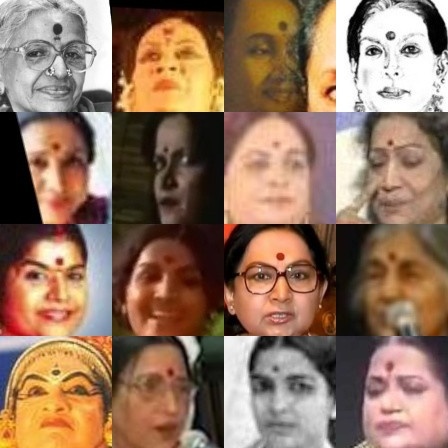}
\captionsetup{justification=centering}
\caption{\textcolor{adairblue}{\textbf{ADA-iR50}}: \(c=\) \emph{forehead dot}\\\(n=5005\)\\\mbox{\(S_n(c)=0.502,\;A_n(c)=\mathrm{0.825}\).}}
\end{subfigure}\hfill
\begin{subfigure}[t]{0.245\textwidth}
\centering
\includegraphics[width=\linewidth]{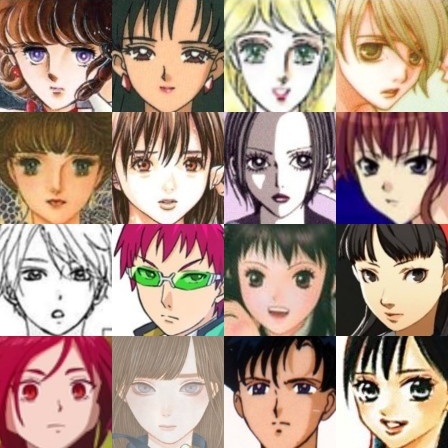}
\captionsetup{justification=centering}
\caption{\textcolor{arcorange}{\textbf{ARC-iR100}}: \(c=\) \emph{animated faces}\\\(n=4709\)\\\mbox{\(S_n(c)=0.768,\;A_n(c)=\mathrm{0.807}\).}}
\end{subfigure}
\vspace{0.1cm}
\caption{Representative open-vocabulary features recovered by SCOUT across different FR backbones. Each block of images shows the top
16 identity-deduplicated activating samples for one recovered feature together with its queried concept \(c\), feature
index \(n\), selectivity \(S_n(c)\), and concept agreement \(A_n(c)\). The reported statistics are computed on the full
retrieved top-activating set. The semantic consistency of these examples indicates that SCOUT can recover meaningful
features beyond a small predefined face-attribute taxonomy.\vspace{-1mm}}
\label{fig:max_activating_samples}
\end{figure*}

\begin{table}[!t]
\centering\caption{FR encoders used in our experiments. The four models span CNN and transformer
backbones with distinct margin-based losses and are trained on different large-scale datasets.\vspace{1mm}}
\label{tab:fr-models}
\small
\setlength{\tabcolsep}{2pt}
\renewcommand{\arraystretch}{1.0}
\resizebox{0.95\columnwidth}{!}{%
\begin{tabular}{@{}
>{\hspace{2pt}\raggedright\arraybackslash}l<{\hspace{2pt}}
>{\hspace{2pt}\raggedright\arraybackslash}l<{\hspace{2pt}}
>{\hspace{2pt}\raggedright\arraybackslash}l<{\hspace{2pt}}
>{\hspace{2pt}\raggedright\arraybackslash}l<{\hspace{2pt}}
@{}}
\toprule
\textbf{Model} & \textbf{Loss} & \textbf{Backbone} & \textbf{Training set} \\
\midrule
\textcolor{adairblue}{\textbf{ADA-iR50}}  & AdaFace~\cite{kim2022adaface} & iResNet-50~\cite{he2016deep} & MS1MV2~\cite{deng2019arcface} \\
\textcolor{arcorange}{\textbf{ARC-iR100}} & ArcFace~\cite{deng2019arcface} & iResNet-100~\cite{he2016deep} & WebFace42M~\cite{zhu2023webface260m} \\
\textcolor{vitgreen}{\textbf{ADA-ViT}}    & AdaFace~\cite{kim2022adaface} & ViT-Base~\cite{dosovitskiy2021image} & WebFace4M~\cite{zhu2023webface260m} \\
\textcolor{swinpurple}{\textbf{SWIN-T}}   & CosFace~\cite{wang2018cosface} & Swin-Tiny~\cite{liu2021swin} & MS1MV2~\cite{deng2019arcface} \\
\bottomrule
\end{tabular}}
\vspace{-2mm}
\end{table}

\subsection{Feature-level Intervention Analysis (Stage 4)}

The final stage validates the semantic hypotheses from Stage~2 by actively changing the
template along individual sparse feature vectors. We refer to such a controlled change
as a \textit{feature-level intervention}. For a selected feature \(n\), let
\(\mathbf{d}_n \in \mathbb{R}^{d}\) denote its template-space dictionary vector.
Using its \(\ell_2\)-normalized form
\(\widetilde{\mathbf{d}}_n = \mathbf{d}_n / \lVert\mathbf{d}_n\rVert\), we define the manipulated template as
\begin{equation}
\mathbf{x}'(\alpha) = (\mathbf{x}+\alpha\,\widetilde{\mathbf{d}}_n) \;/\; \lVert\mathbf{x}+\alpha\,\widetilde{\mathbf{d}}
_n\rVert,
\end{equation}
where $\alpha \in \mathbb{R}$ controls the intervention strength. Normalizing $\mathbf{d}
_n$ makes $\alpha$ comparable across features, while re-projecting keeps $
\mathbf{x}'(\alpha)$ on the unit hypersphere.

To inspect the effect of the intervention in image space, we reconstruct $I'(\alpha)$ from
$\mathbf{x}'(\alpha)$ using the inversion models from Stage~3. The interventions considered
here are local and asymmetric: positive traversal ($\alpha>0$) amplifies the selected
feature, whereas negative traversal ($\alpha<0$) is not assumed to realize its semantic
opposite. We quantify the effect through two complementary metrics that capture semantic
control and identity preservation.

\vspace{0.5mm}\noindent\textbf{Attribute Gain ($\Delta_c$).}
We quantify semantic amplification with the evaluation confidence score $s_c^{\mathrm{eval}}$, computed with the reserved evaluation prompt bank,
\begin{equation}
\Delta_c(\alpha)=s_c^{\mathrm{eval}}(I'(\alpha)) - s_c^{\mathrm{eval}}(I'(0)).
\end{equation}
Higher positive values indicate stronger expression of the queried concept, whereas values
near zero or below indicate little or no amplification.

\vspace{0.5mm}\noindent\textbf{Template Similarity ($\theta_c$).}
We measure the displacement between the manipulated and original
templates using cosine similarity,
\begin{equation}
\theta_c(\alpha) = \cos(\mathbf{x}'(\alpha),\, \mathbf{x}).
\end{equation}
This provides a descriptive measure of template displacement.
Fixed-FMR identity evaluation is performed separately using FNMR at a
frozen verification threshold.

\begin{table}[!t]
\centering
\caption{SAE statistics on Glint360K ($\sim17$M templates) for the four SCOUT models, alongside FaceMINT~\cite{rot2025facemint} TopK SAE reference results. MSE $=\mathbb{E}[\lVert\mathbf{x}-\hat{\mathbf{x}}\rVert^2]$, alive\% is the fraction of features activated at least once, and $|\cos|_{\mathrm{dict}}$ is the mean pairwise similarity between template-space dictionary vectors.\vspace{1mm}}
\label{tab:sae-training}
\scriptsize
\setlength{\tabcolsep}{2.5pt}
\renewcommand{\arraystretch}{1.2}
\resizebox{\columnwidth}{!}{%
\begin{tabular}{@{}>{\centering\arraybackslash}m{0.82cm}
>{\raggedright\arraybackslash}m{1.5cm}
c c c c c c@{}}
\toprule
\textbf{Method} &
\textbf{Backbone} &
$\boldsymbol{d_{\mathrm{SAE}}}$ &
$\boldsymbol{k}$ &
\textbf{MSE} &
\textbf{alive\%} &
$\boldsymbol{\cos(\mathbf{x},\hat{\mathbf{x}})}$ &
$\boldsymbol{|\cos|_{\mathrm{dict}}}$ \\
\midrule

\multirow{4}{0.82cm}{\centering\rotatebox[origin=c]{90}{\textit{SCOUT}}}
& \textcolor{adairblue}{\textbf{ADA-iR50}}
& $8192$ & $32$ & $0.0008$ & $100.00$ & $0.7743$ & $0.0465$ \\
& \textcolor{vitgreen}{\textbf{ADA-ViT}}
& $8192$ & $32$ & $0.0007$ & $100.00$ & $0.8099$ & $0.0547$ \\
& \textcolor{arcorange}{\textbf{ARC-iR100}}
& $8192$ & $32$ & $0.0008$ & $100.00$ & $0.7694$ & $0.0450$ \\
& \textcolor{swinpurple}{\textbf{SWIN-T}}
& $8192$ & $32$ & $0.0006$ & $100.00$ & $0.8284$ & $0.0579$ \\

\midrule

\multirow{2}{0.82cm}{\centering\rotatebox[origin=c]{90}{\textit{\shortstack{Face\\MINT}}}}
& \textcolor{adairblue}{\textbf{ADA-iR50}}
& $65536$ & $400$ & $0.0003$ & $99.82$ & $0.9998$ & $0.0767$ \\
& \textcolor{swinpurple}{\textbf{SWIN-T}}
& $65536$ & $200$ & $0.0000$ & $88.15$ & $1.0000$ & $0.0365$ \\

\bottomrule
\end{tabular}}
\vspace{-2mm}
\end{table}

\section{Experiments}

This section describes the experimental setting used to evaluate SCOUT.~We first introduce the datasets, FR backbones, and concept-scoring setup used throughout the study, then summarize key implementation details and the evaluation protocol.

\subsection{Experimental Setup}

\noindent\textbf{Datasets.}~We use Glint360K~\cite{an2021partial} for Stage~1 SAE training and Stage~2 concept
discovery, including selectivity ranking and concept-agreement computation. For downstream
inversion-based inspection and feature-level intervention analysis, we use
LFW~\cite{huang2008labeled}. For comparison against image-space editing baselines, we use
CelebA-HQ~\cite{karras2018progressive}.

\vspace{0.5mm}\noindent\textbf{Face Recognition Templates.}~We evaluate SCOUT on four representative FR encoders spanning CNN and transformer
backbones, distinct margin-based training objectives, and different training data sources,
as summarized in Table~\ref{tab:fr-models}. All models operate on aligned $112\times112$
face crops and produce $\ell_2$-normalized 512-d templates, enabling a controlled
assessment of whether the pipeline transfers across heterogeneous template geometries rather than being tied to a single FR model.

\begin{figure*}[!t]
\centering

\begin{minipage}[c]{0.7\textwidth}
\centering

\resizebox{\linewidth}{!}{%
\arrayrulecolor{black!35}%
\begin{tabular}{
    C{2.00cm}C{1.64cm}@{}C{1.64cm}@{}C{1.64cm}
    @{\hspace{0.12cm}}:
    C{2.00cm}C{1.64cm}@{}C{1.64cm}@{}C{1.64cm}
}
\inputimagelabel
  & \alphacell{\alpha=0.00} & \alphacell{\alpha=0.10} & \alphacell{\alpha=0.20}
  & \inputimagelabel & \alphacell{\alpha=0.00} & \alphacell{\alpha=0.20} & \alphacell{\alpha=0.40}\\[\rowskip]

\targetupper{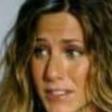}
  & \imgcellfile{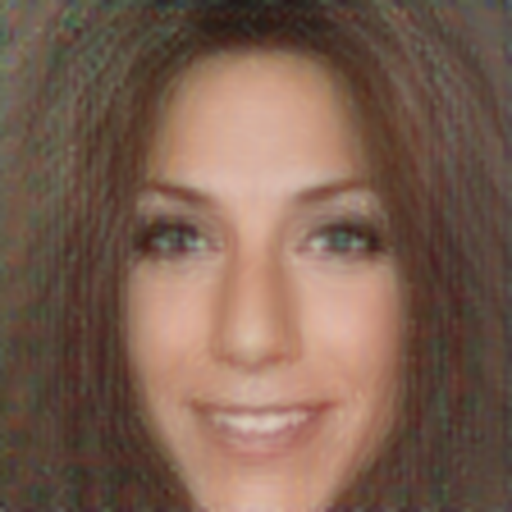}
  & \imgcellfile{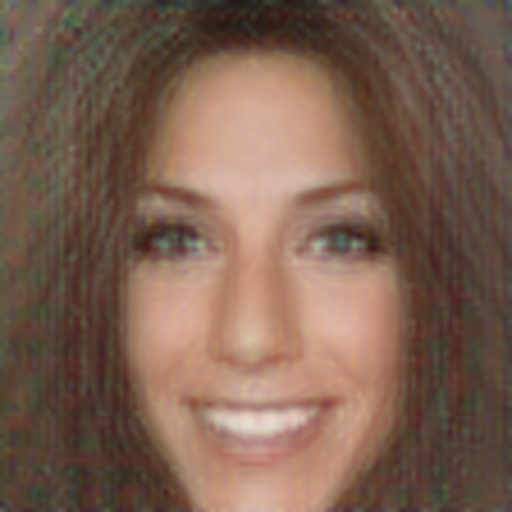}
  & \imgcellfile{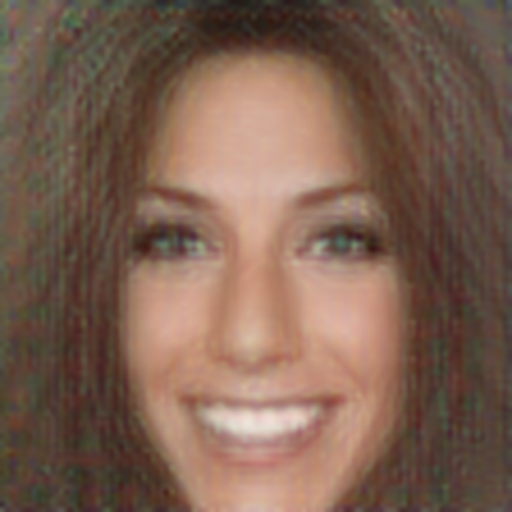}
  & \targetupper{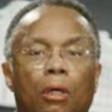}
  & \imgcellfile{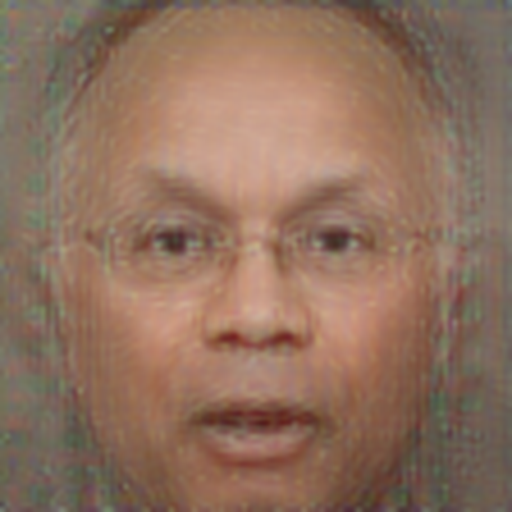}
  & \imgcellfile{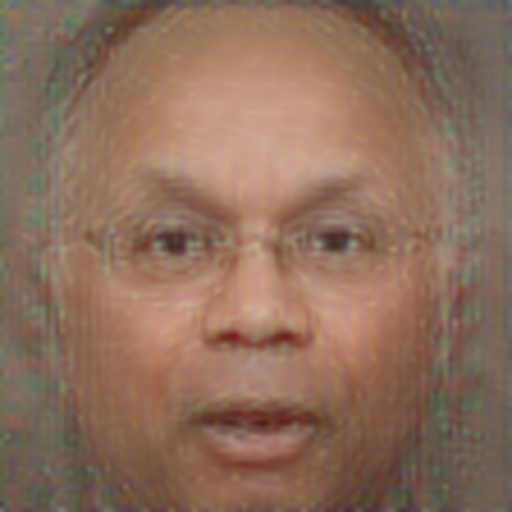}
  & \imgcellfile{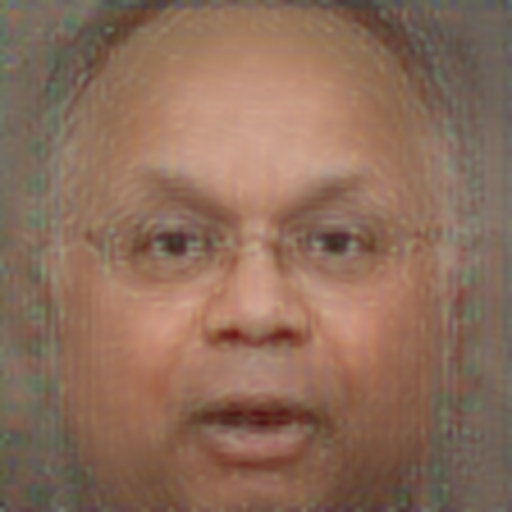}\\[\rowskip]

\multirow{3}{*}{\targetlower{smiling}{4656}{0.249}{83.9\%}}
  & \deltacell{0.000} & \deltacell{0.215} & \deltacell{0.320}
  & \multirow{3}{*}{\targetlower{overweight}{4463}{0.554}{89.5\%}}
  & \deltacell{0.000} & \deltacell{0.048} & \deltacell{0.277}\\

  & \imgcellfile{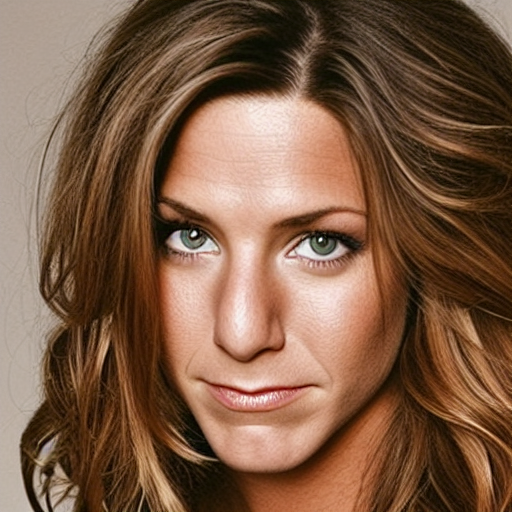}
  & \imgcellfile{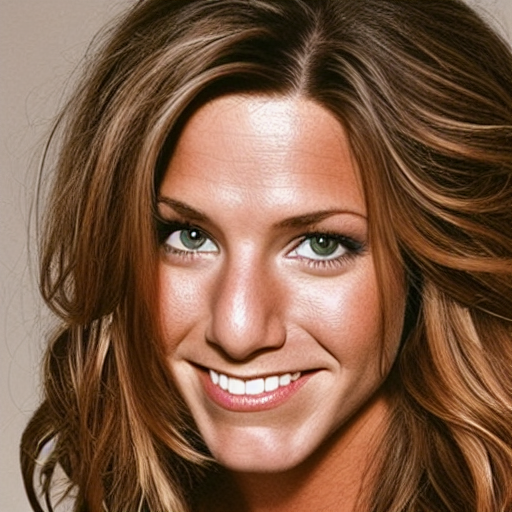}
  & \imgcellfile{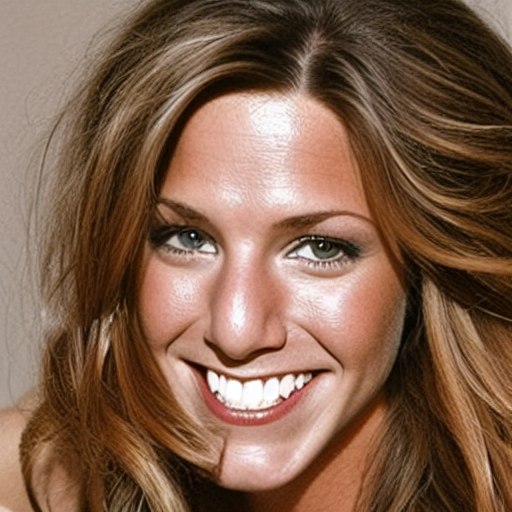}
  &
  & \imgcellfile{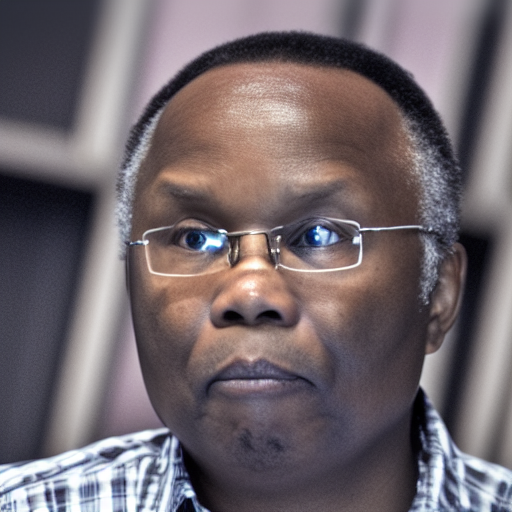}
  & \imgcellfile{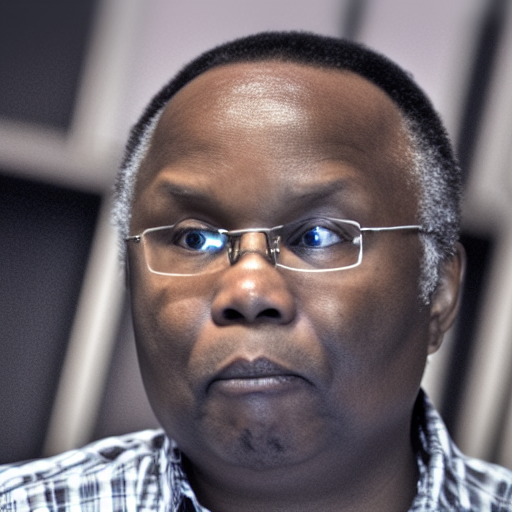}
  & \imgcellfile{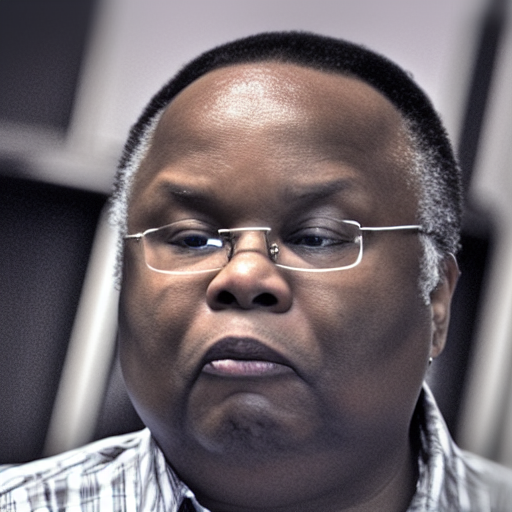}\\[\rowskip]

  & \deltacell{0.000} & \deltacell{0.726} & \deltacell{0.943}
  &
  & \deltacell{0.000} & \deltacell{0.307} & \deltacell{0.436}\\

\hdashline
\rowcolor{metricgray}
\targetcossim & \cossimcell{1.000} & \cossimcell{0.995} & \cossimcell{0.981}
  & \targetcossim & \cossimcell{1.000} & \cossimcell{0.980} & \cossimcell{0.928}\\
\hdashline
\end{tabular}%
\arrayrulecolor{black}%
}

\vspace{1pt}

\resizebox{\linewidth}{!}{%
\arrayrulecolor{black!35}%
\begin{tabular}{
    C{2.00cm}C{1.64cm}@{}C{1.64cm}@{}C{1.64cm}
    @{\hspace{0.12cm}}:
    C{2.00cm}C{1.64cm}@{}C{1.64cm}@{}C{1.64cm}
}
\inputimagelabel
  & \alphacell{\alpha=0.00} & \alphacell{\alpha=0.40} & \alphacell{\alpha=0.60}
  & \inputimagelabel & \alphacell{\alpha=0.00} & \alphacell{\alpha=0.15} & \alphacell{\alpha=0.30}\\[\rowskip]

\targetupper{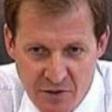}
  & \imgcellfile{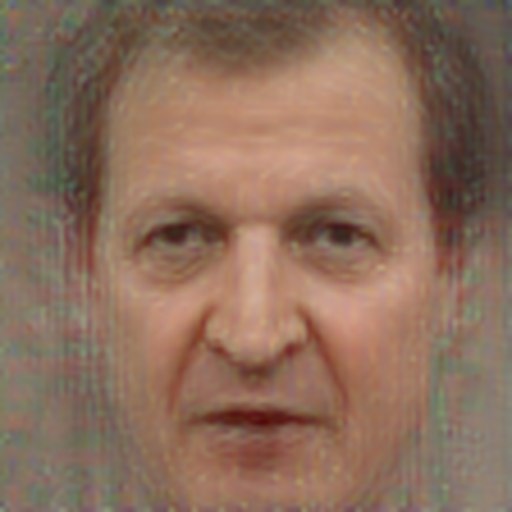}
  & \imgcellfile{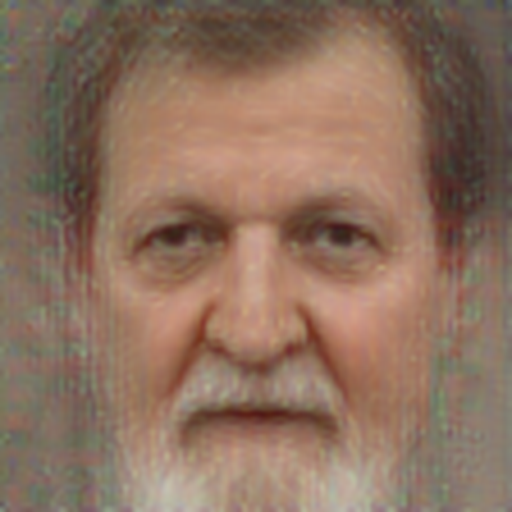}
  & \imgcellfile{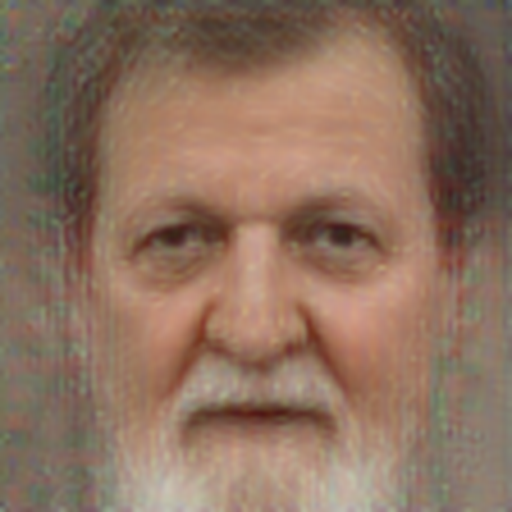}
  & \targetupper{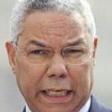}
  & \imgcellfile{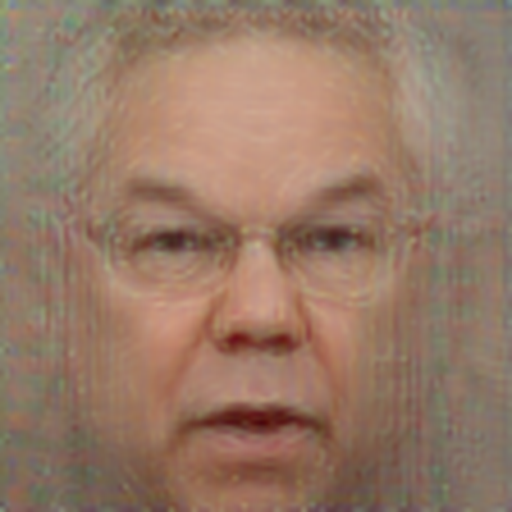}
  & \imgcellfile{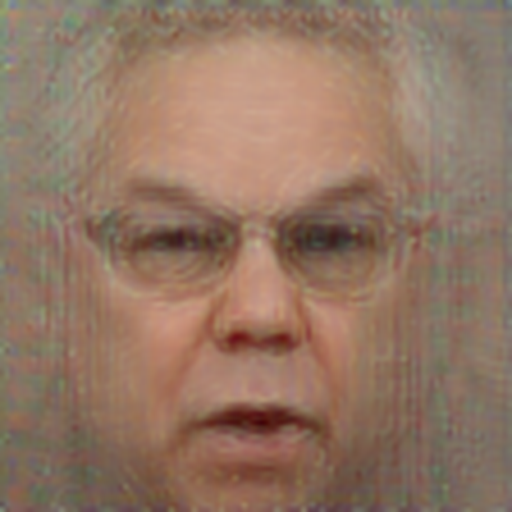}
  & \imgcellfile{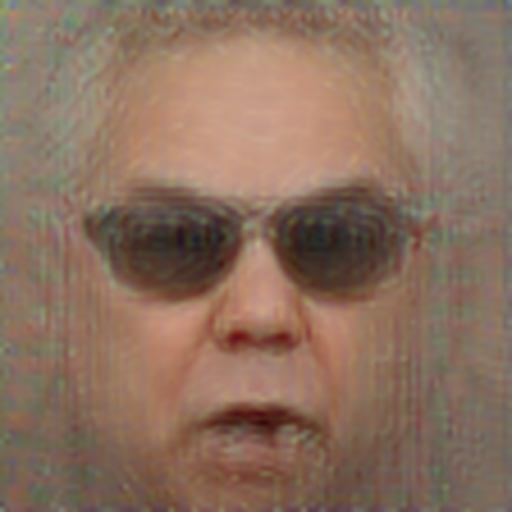}\\[\rowskip]

\multirow{3}{*}{\targetlower{gray beard}{3507}{0.468}{80.0\%}}
  & \deltacell{0.000} & \deltacell{0.799} & \deltacell{0.805}
  & \multirow{3}{*}{\targetlower{sunglasses}{8020}{0.356}{89.5\%}}
  & \deltacell{0.000} & \deltacell{0.052} & \deltacell{0.359}\\

  & \imgcellfile{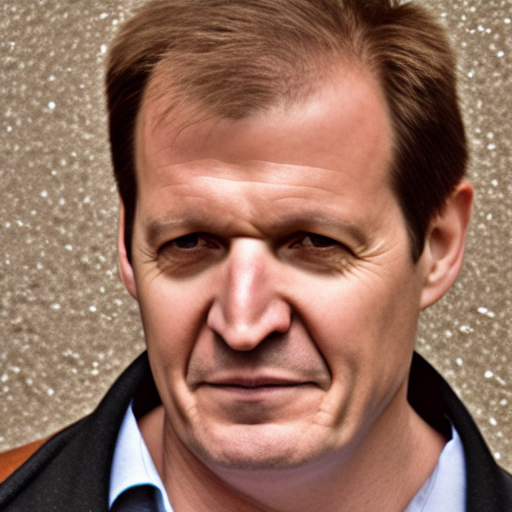}
  & \imgcellfile{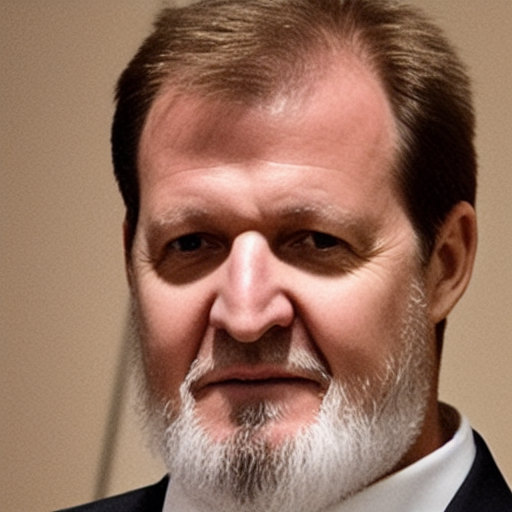}
  & \imgcellfile{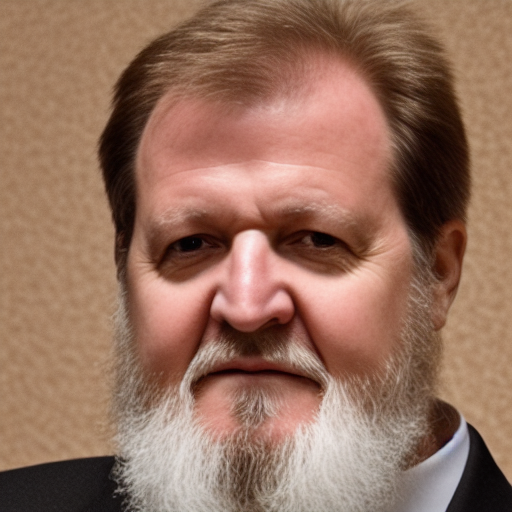}
  &
  & \imgcellfile{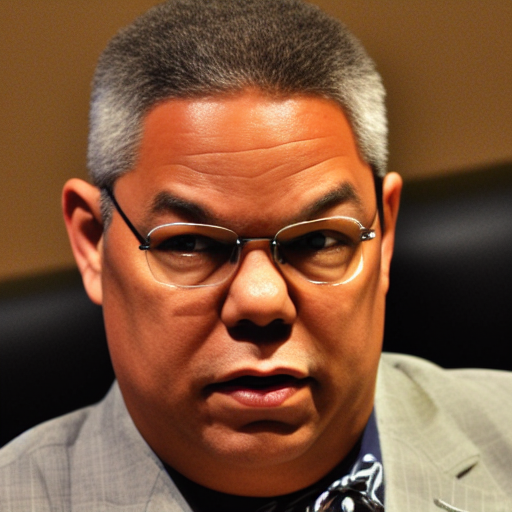}
  & \imgcellfile{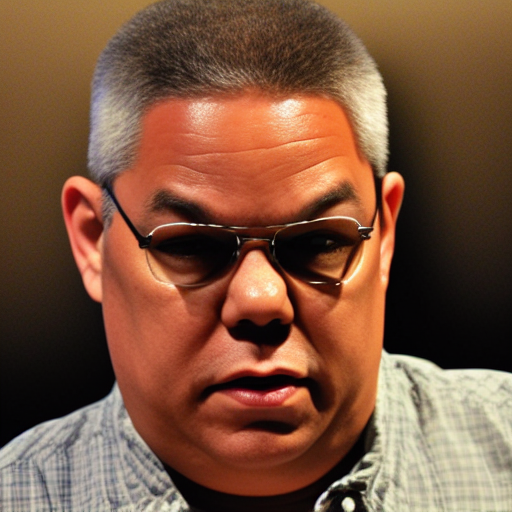}
  & \imgcellfile{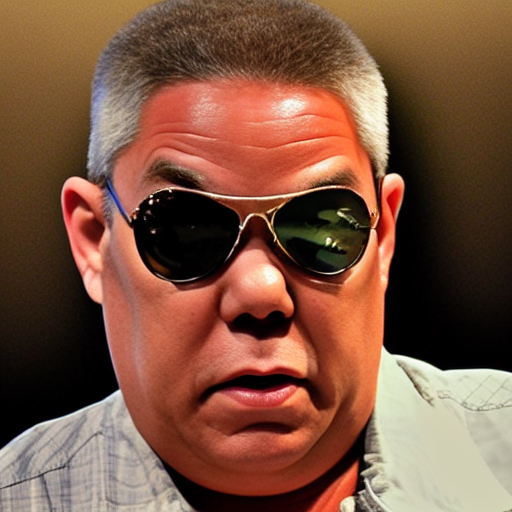}\\[\rowskip]

  & \deltacell{0.000} & \deltacell{0.665} & \deltacell{0.682}
  &
  & \deltacell{0.000} & \deltacell{0.248} & \deltacell{0.273}\\

\hdashline
\rowcolor{metricgray}
\targetcossim & \cossimcell{1.000} & \cossimcell{0.929} & \cossimcell{0.860}
  & \targetcossim & \cossimcell{1.000} & \cossimcell{0.989} & \cossimcell{0.957}\\
\hdashline
\end{tabular}%
\arrayrulecolor{black}%
}

\end{minipage}%
\hfill%
\begin{minipage}[c]{0.26\textwidth}

\hspace*{-0.75cm}\includegraphics[width=1.2\textwidth]{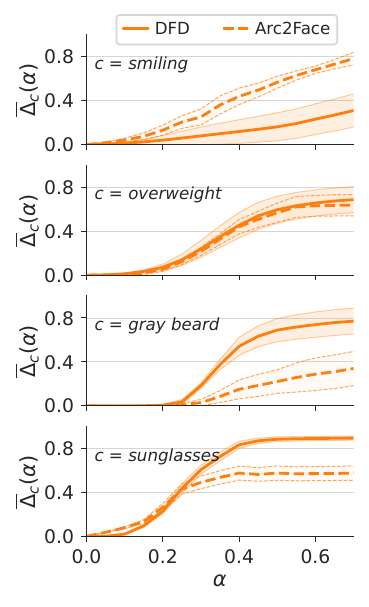}
\end{minipage}

\vspace{0.1cm}
\caption{Cross-decoder validation in the shared ARC-iR100 space. Left: one
intervened template per concept is decoded by DFD and Arc2Face;
\(\Delta_c\) is averaged across the disjoint evaluation-prompt subsets, while
the shared template-level \(\theta_c\) is shown once per intervention
strength. Metadata \(n\), \(S_n(c)\), and \(A_n(c)\) are obtained in Stage~2. Right: attribute-gain curves over 100 random LFW templates. Thick solid and
dashed lines denote the DFD and Arc2Face means across evaluation-prompt subsets, respectively;
thin matching boundaries and shading denote the \(\pm1\) standard deviation across evaluation-prompt subsets.}
\label{fig:qualitative_grid}\vspace{-5mm}
\end{figure*}

\vspace{0.5mm}\noindent\textbf{Concept Scoring.}~We instantiate separate discovery and evaluation scores with CLIP ViT-L/14~\cite{radford2021learning}. Stage~2 uses three positive and three negative discovery prompts to obtain $s_c^{\mathrm{disc}}$ for exemplar construction and feature selection. After the direction is fixed, Stage~4 uses disjoint pools of five positive and five negative evaluation prompts, with no overlap with the discovery prompts. We evaluate all 100 possible 3-vs.-3 subsets and report the mean and pointwise $\pm1$ standard deviation across the resulting population curves. This protocol additionally evaluates robustness to evaluation-prompt wording.

\begin{table}[!t]
\centering
\caption{Fixed-FMR identity evaluation on LFW: FNMR (\%) at a threshold calibrated on clean
impostor pairs to FMR$=10^{-2}$ and frozen after manipulation. Perturbed
statistics summarize the full SAE-direction sweep.}\vspace{0.5mm}
{
\footnotesize
\setlength{\tabcolsep}{4pt}
\renewcommand{\arraystretch}{1.03}
\begin{tabular*}{\linewidth}{@{\extracolsep{\fill}}lrrrr@{}}
\toprule
& &
\multicolumn{3}{c}{\textbf{Perturbed ($\alpha=1$)}} \\
\cmidrule(l){3-5}
\textbf{Model} & \textbf{Clean} & \textbf{Mean} & \textbf{p95} & \textbf{Worst} \\
\midrule
ADA-iR50  & $0.33$ & $0.43$ & $0.53$ & $0.67$ \\
ADA-ViT   & $0.23$ & $0.41$ & $0.53$ & $0.73$ \\
SWIN-T    & $0.20$ & $0.41$ & $0.50$ & $0.67$ \\
ARC-iR100 & $0.27$ & $0.42$ & $0.50$ & $0.63$ \\
\bottomrule
\end{tabular*}\vspace{-3mm}

\label{tab:lfw-fixed-fmr}
}\vspace{-3mm}
\end{table}

\subsection{Implementation Details}

\noindent\textbf{Concept-Discovery Settings.}~In all concept-discovery experiments, we use \(|\mathcal{D}^{+}_c|=|\mathcal{D}^{-}_c|=1{,}
024\).~Features are ranked by \(S_n(c)\), the top~10 are retained, and concept agreement is
computed on the top \(|\mathcal{T}_n|=10{,}000\) identity-deduplicated activating samples.
We keep only the features satisfying \(A_n(c)\geq 80\%\) as final candidates.

For SCOUT, we use \(d_{\mathrm{SAE}}{=}8{,}192\) and mean sparsity \(k{=}32\).~To prevent feature collapse, we include the auxiliary dead-latent reconstruction loss of Gao \etal~\cite{gao2024scaling}. Across all four encoders, our SAEs achieve low
reconstruction error (in terms of MSE), 100\% alive features, and low mean pairwise dictionary-vector similarity, indicating a diverse, non-redundant feature set.

Table~\ref{tab:sae-training} summarizes reconstruction quality and reports FaceMINT~\cite{rot2025facemint} TopK SAE results
under the same metrics, included as the conceptually closest baseline to our BatchTopK configuration. Compared with FaceMINT, SCOUT
has lower reconstruction cosine similarity, which is expected given our substantially more compact SAE configuration.
In SCOUT, the SAE serves primarily as a sparse feature dictionary for concept discovery, rather than as a
high-fidelity reconstruction model. We therefore favor a compact configuration that supports efficient large-scale concept
search while still yielding stable, diverse feature directions.
  
\vspace{0.5mm}\noindent\textbf{Computational Complexity.}~All experiments were run on a single NVIDIA RTX~3090. The one-time costs per FR encoder are SAE training on Glint360K (\(\approx 20\) min) and DFD training (\(\approx 1\) day). After the training, the full per-concept SCOUT pipeline, from concept scoring through intervention-based evaluation, completes in under five minutes per concept.

\begin{figure*}[!t]
\begingroup
\setlength{\tabcolsep}{0pt}
\newlength{\qualimgwd}\setlength{\qualimgwd}{1.18cm}
\newlength{\quallabelwd}\setlength{\quallabelwd}{0.38cm}
\newlength{\qualgridwd}\setlength{\qualgridwd}{\dimexpr\quallabelwd+8\qualimgwd\relax}
\newcommand{\qualclass}[1]{\parbox[c][0.24cm][c]{\dimexpr2\qualimgwd\relax}{\centering\scriptsize #1}}
\newcommand{\qualmethod}[1]{\parbox[c][\qualimgwd][c]{\quallabelwd}{\centering\rotatebox[origin=c]{90}{\scriptsize\bfseries\strut #1}}}
\newcommand{\qualimg}[1]{\parbox[c][\qualimgwd][c]{\qualimgwd}{\includegraphics[width=\qualimgwd,height=\qualimgwd,trim=0.7pt 0.7pt 0.7pt 0.7pt,clip]{#1}}}
\newcommand{\qualdelta}[1]{\parbox[c][0.22cm][c]{\qualimgwd}{\centering\tiny $\overline\Delta_c{=}#1$}}
\newcommand{\qualhead}{\makebox[\qualgridwd][l]{\makebox[\quallabelwd][c]{}\qualclass{Bald}\qualclass{Smiling}\qualclass{Eyeglasses}\qualclass{East Asian}}}
\newcommand{\qualfoot}{\makebox[\qualgridwd][l]{\makebox[\quallabelwd][c]{}\qualdelta{0}\qualdelta{0.6}\qualdelta{0}\qualdelta{0.6}\qualdelta{0}\qualdelta{0.6}\qualdelta{0}\qualdelta{0.4}}}
\newcommand{\qualrow}[9]{\makebox[\qualgridwd][l]{\qualmethod{#1}\qualimg{#2}\qualimg{#3}\qualimg{#4}\qualimg{#5}\qualimg{#6}\qualimg{#7}\qualimg{#8}\qualimg{#9}}}

\noindent\makebox[\textwidth][c]{%
\begin{minipage}[c]{0.430\textwidth}
\vspace{0pt}\centering
\includegraphics[width=\linewidth]{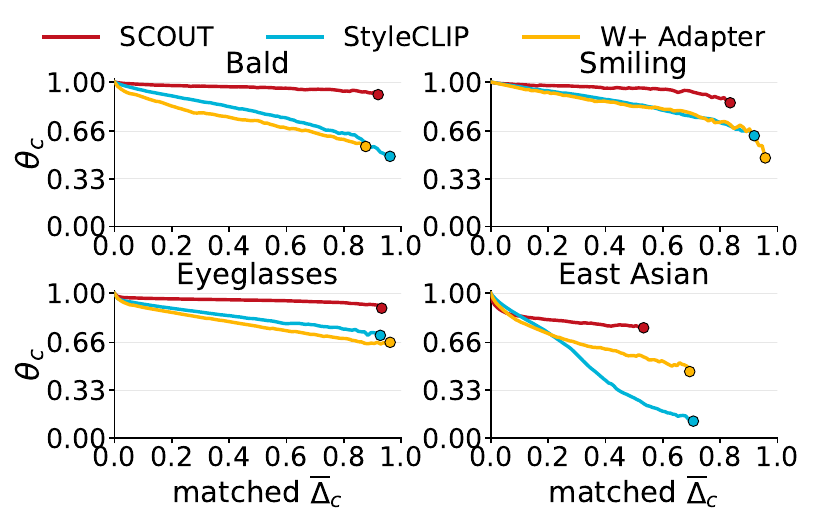}
\end{minipage}\hfill
\begin{minipage}[c]{0.585\textwidth}
\vspace{0pt}\centering\offinterlineskip
\qualhead\par
\qualrow{SCOUT}
 {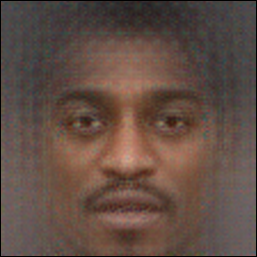}{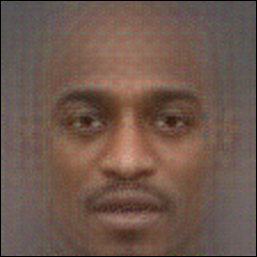}
 {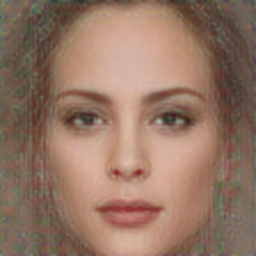}{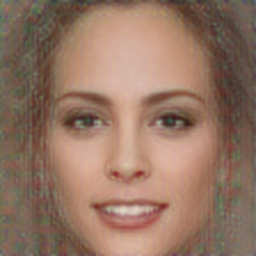}
 {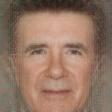}{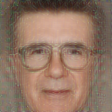}
 {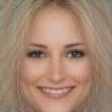}{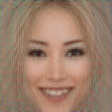}\par
\qualrow{StyleCLIP}
 {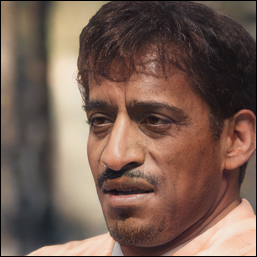}{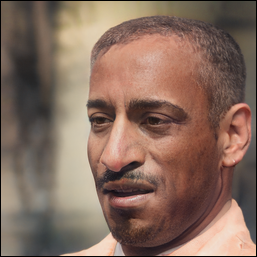}
 {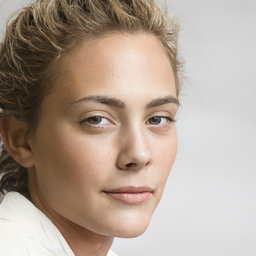}{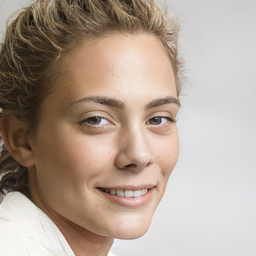}
 {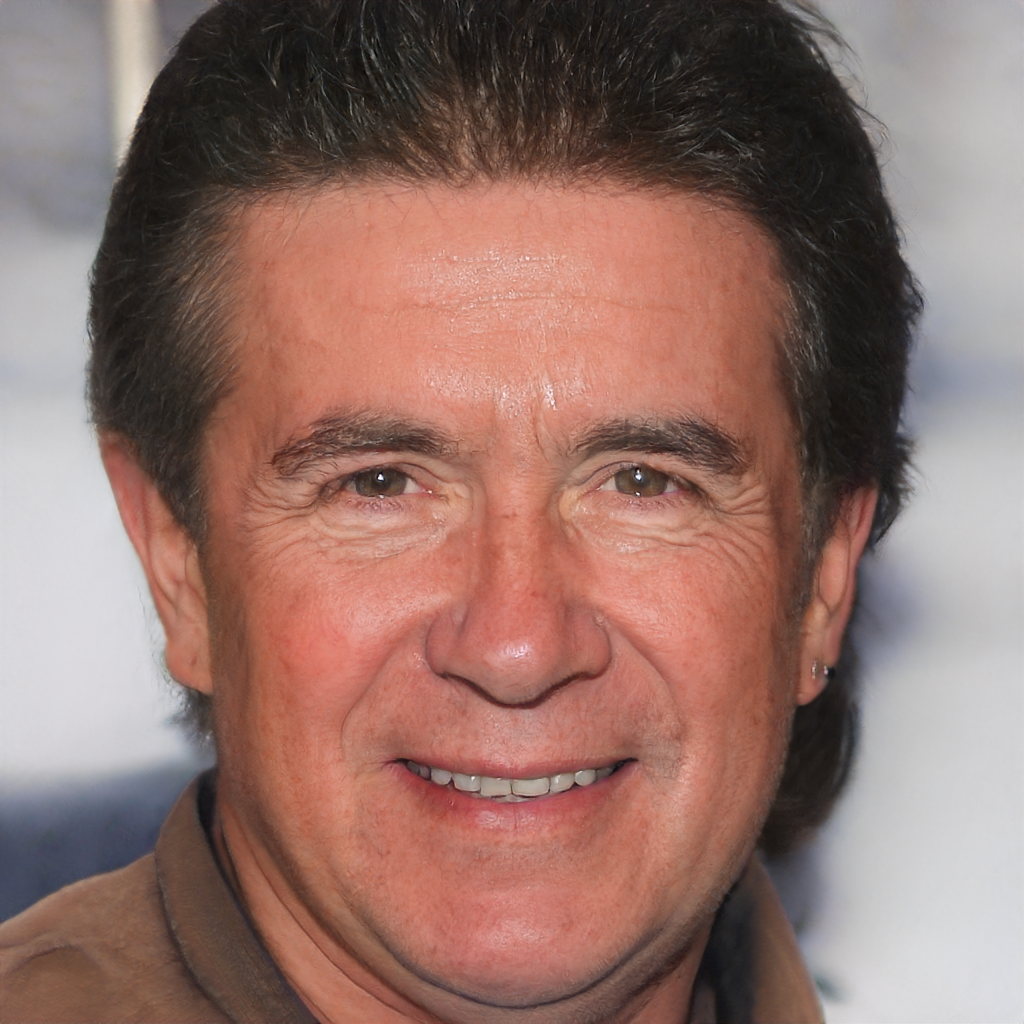}{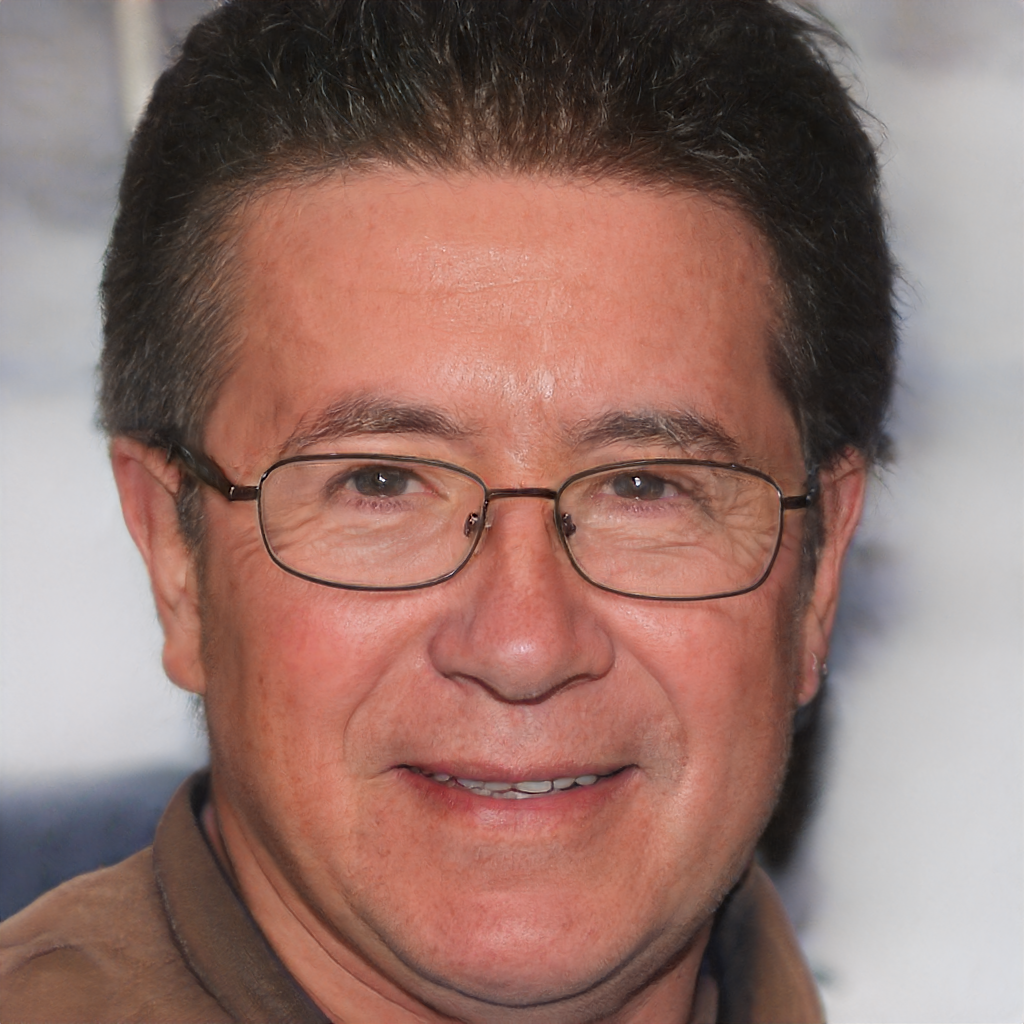}
 {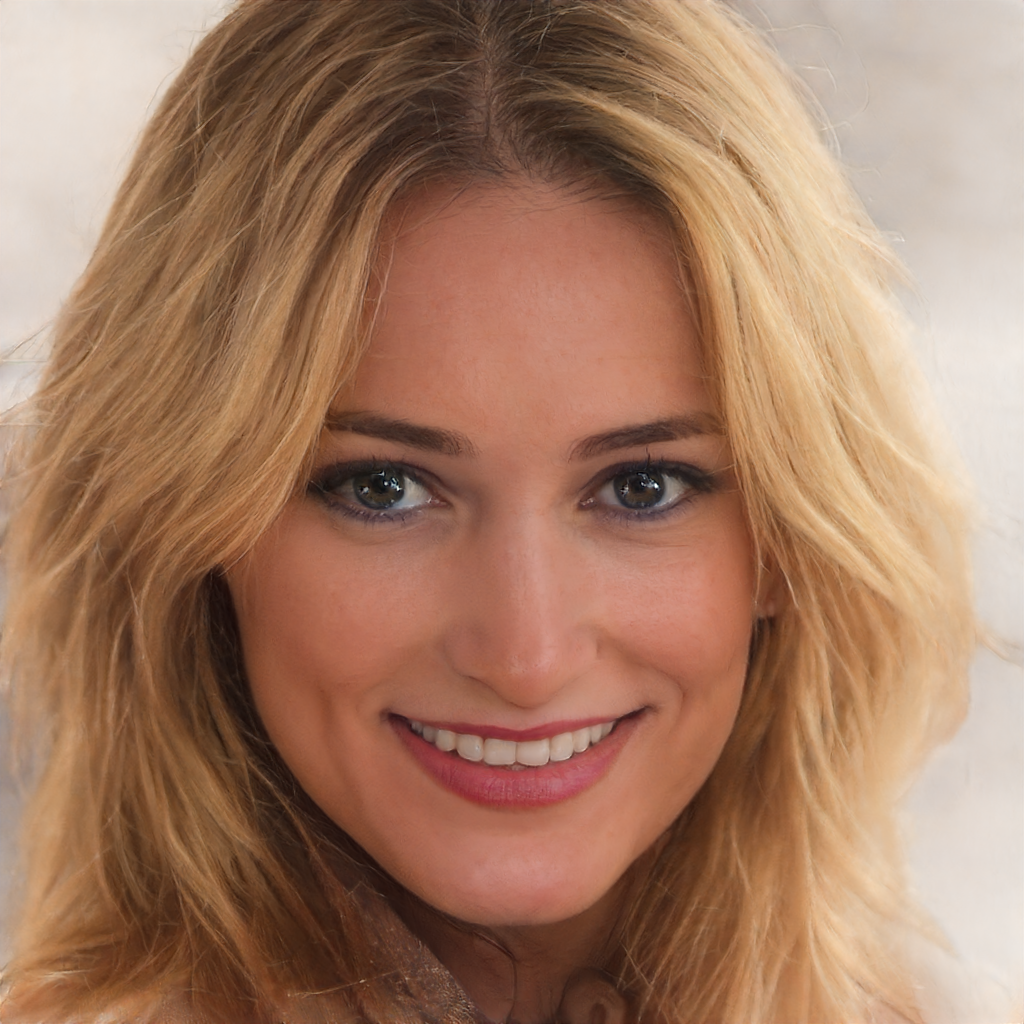}{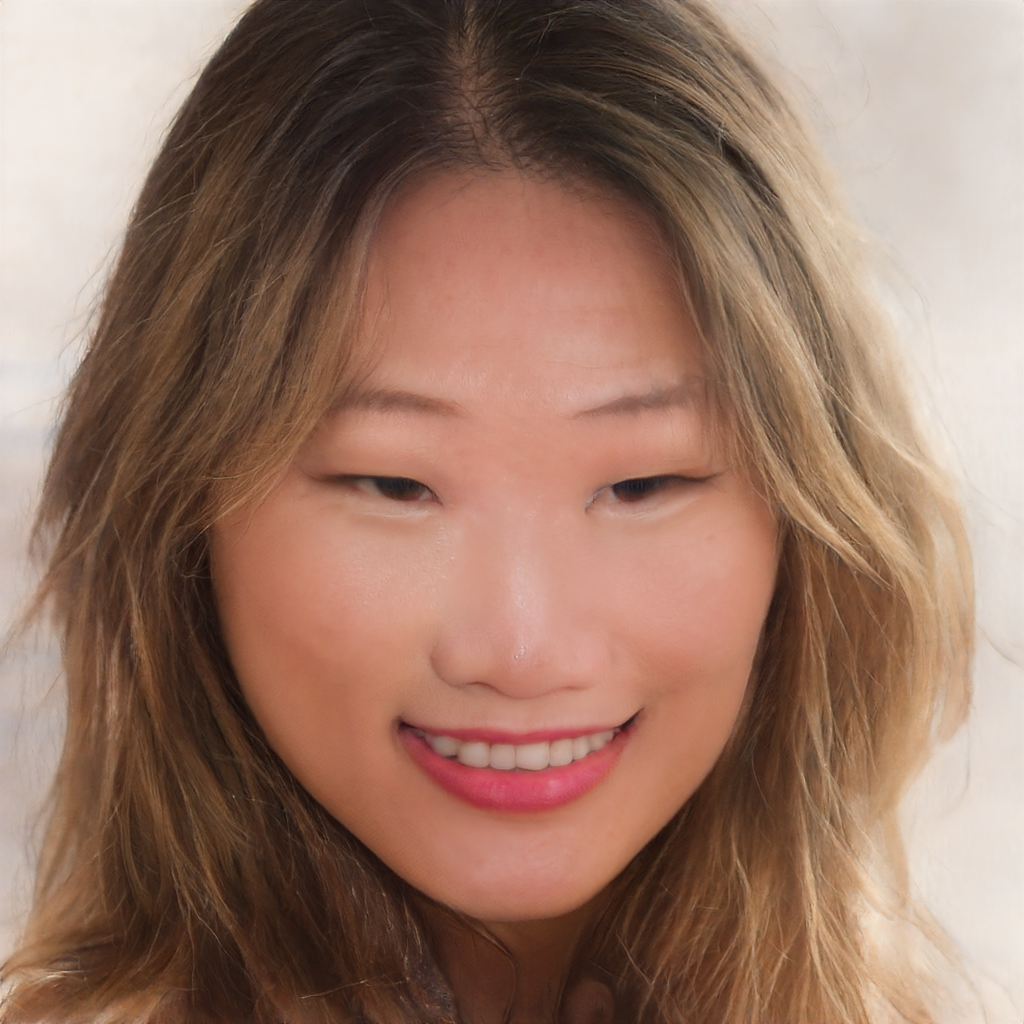}\par
\qualrow{W+}
 {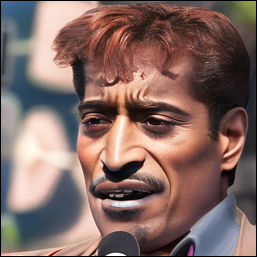}{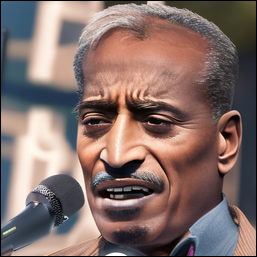}
 {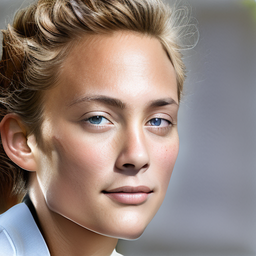}{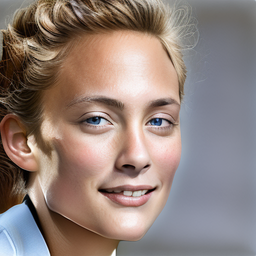}
 {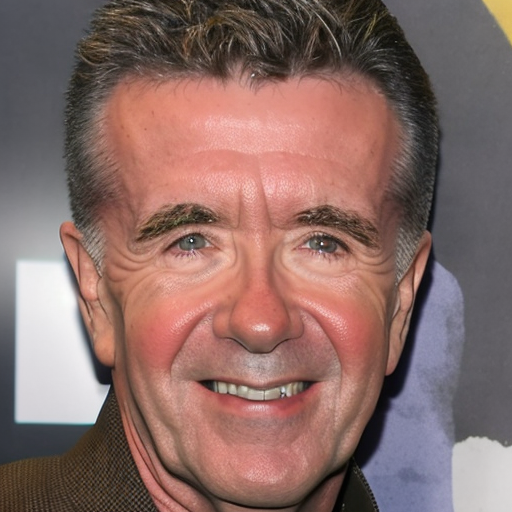}{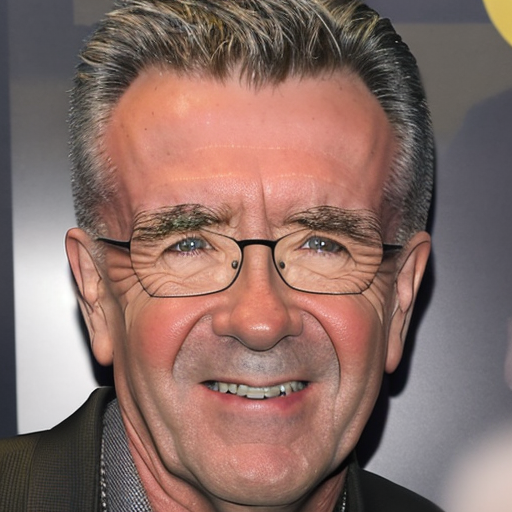}
 {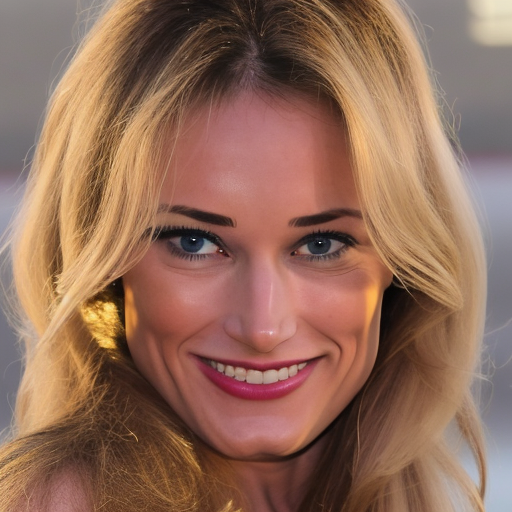}{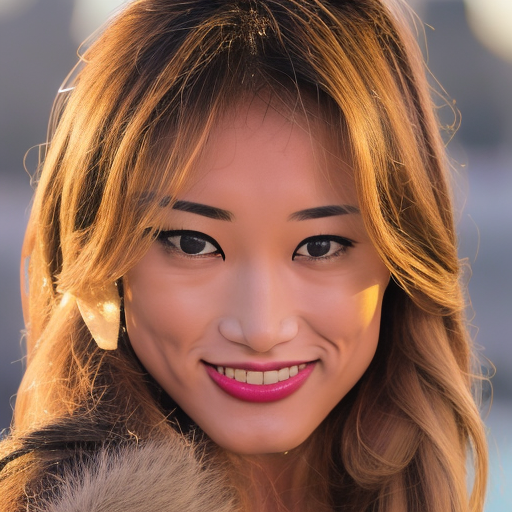}\par
\qualfoot
\end{minipage}}
\endgroup

\vspace{0.10cm}
\caption{Identity preservation at matched mean attribute gain on CelebA-HQ (100 samples per attribute). Left: $\theta_c$ versus $\overline\Delta_c$ for SCOUT, StyleCLIP, and W\(+\) Adapter; dots mark each method's maximum population-supported $(\overline\Delta_{\max},\theta)$, reported per concept in Section~\ref{sec:results}. Right: representative matched-gain edits; target gains are shown below each image.}
\label{fig:celebahq_method_comparison}
\end{figure*}

\subsection{Evaluation Protocol}\label{subsec:eval-protocol}

\noindent\textbf{Primary Case Study.}~For the main end-to-end demonstration, the Stage-2 discovery bank for \(c=\) \emph{bald} contains positive prompts
\{``bald person'', ``hairless head'', ``visible scalp''\} and negative prompts \{``thick hair'', ``voluminous hair'',
``hair parting''\}. The other discovery banks are specified analogously. Stage-4 semantic-gain curves follow the disjoint evaluation-prompt protocol above and use the same 100 random LFW templates. Their lines and bands report the mean and pointwise \(\pm1\) standard deviation across evaluation-prompt subsets.

\vspace{0.5mm}\noindent\textbf{Fixed-FMR Identity Evaluation.}~We sweep every SAE direction at the strong intervention
strength $\alpha=1$, yielding a conservative, worst-case estimate. For each LFW pair, one template is manipulated while
the other remains fixed. For each FR model, we calibrate a verification
threshold on clean impostor pairs at FMR$=10^{-2}$, freeze it, and reuse it
after manipulation. We report clean FNMR and the mean, 95th-percentile, and
worst perturbed FNMR across directions.

\vspace{0.5mm}\noindent\textbf{Comparison to Image-Space Editing Baselines.}~We compare SCOUT to two image-space editing baselines, StyleCLIP~\cite{patashnik2021styleclip} and W\(+\)
Adapter~\cite{li2024wplusadapter}.~Because both baselines operate on inverted high-quality face images, we use 100 randomly sampled CelebA-HQ~\cite{karras2018progressive} images per attribute. All curves include all four FR backbones. For all three methods, we sweep the manipulation strength and compare identity preservation at the same mean attribute gain. For each image--backbone pair, we retain the increasing branch of the gain--identity trace up to its individual maximum. At each gain, we average template similarity over the traces that attain it. Curves terminate when fewer than 10\% of the pooled traces support the gain. The endpoint therefore represents the maximum population-supported gain. For SCOUT, \(\theta_c\) is measured directly between the original
and manipulated templates, while \(\overline\Delta_c\) is computed from the corresponding decodings. For StyleCLIP and W\(+\) Adapter,
both metrics are computed after image editing and re-encoding.

\vspace{0.5mm}\noindent\textbf{Cross-Decoder Validation.}~To assess whether the recovered directions remain semantically coherent across inversion backends, we decode the same
template-space interventions for four additional concepts with both DFD and Arc2Face. These comparisons use only
ARC-iR100, since Arc2Face is conditioned on ArcFace embeddings.

\section{Results}\label{sec:results}

\noindent\textbf{Full Pipeline Demonstration.}~We first trace the queried concept \(c=\) \emph{bald} through the full SCOUT pipeline.
Figure~\ref{fig:qualitative_backbones} reports the selected feature for each backbone together with its index \(n\),
selectivity \(S_n(c)\), and concept agreement \(A_n(c)\), revealing strong alignment with the query.~Under disjoint evaluation-prompt scoring, mean baldness gain increases with \(\alpha\) across all four FR backbones. At \(\alpha=0.4\), \(\overline{\Delta}_c\) ranges from \(0.283\) to \(0.527\). The bands quantify variation across evaluation-prompt subsets, while template similarity remains largely preserved.

\vspace{0.5mm}\noindent\textbf{Open-Vocabulary Concept Discovery.}~Beyond the primary case study, SCOUT also recovers coherent concept-linked sparse features for broader free-form queries.
Figure~\ref{fig:max_activating_samples} shows representative examples together with the corresponding feature index \(n\),
selectivity \(S_n(c)\), and concept agreement \(A_n(c)\). The identity-deduplicated top activations remain semantically
consistent within each recovered feature, including concepts that would be difficult to capture with a small predefined
face-attribute taxonomy. This demonstrates the utility of open-vocabulary querying for discovering
semantically meaningful sparse features in settings where no predefined attribute vocabulary is available.

\vspace{0.5mm}\noindent\textbf{Cross-Decoder Validation.}~Figure~\ref{fig:qualitative_grid} shows that the DFD and Arc2Face mean curves move in the same semantic direction for all four concepts, although the gain magnitude varies between decoders. Variation across evaluation-prompt subsets does not alter the overall population-level trends, making a decoder-specific explanation less likely.

\vspace{0.5mm}\noindent\textbf{Fixed-FMR Identity Evaluation.}~
Table~\ref{tab:lfw-fixed-fmr} shows that, even at $\alpha=1$,
the most disruptive direction increases FNMR by at most $0.50$ percentage
points, i.e., verification behavior at the frozen operating
point remains essentially unchanged.

Figure~\ref{fig:celebahq_method_comparison} extends the comparison to four concepts and reports each method's maximum population-supported gain. SCOUT reaches strong gains while preserving identity across all four concepts. For bald and smiling, SCOUT attains $(\overline{\Delta}_{\max},\theta)$ of $(0.92,0.91)$ and $(0.84,0.86)$, with markedly higher identity preservation than StyleCLIP, $(0.96,0.49)$ and $(0.92,0.63)$, and W\(+\) Adapter, $(0.88,0.55)$ and $(0.96,0.47)$. For eyeglasses, SCOUT reaches $(0.93,0.90)$, compared with StyleCLIP $(0.93,0.71)$ and W\(+\) Adapter $(0.96,0.66)$. For East Asian, image-space methods attain higher maximum gain but with substantially lower identity preservation: StyleCLIP reaches $(0.71,0.12)$ and W\(+\) Adapter $(0.69,0.46)$, while SCOUT reaches $(0.53,0.76)$. Thus, direct template-level intervention offers a stronger fidelity--edit trade-off. This comparison is conservative for StyleCLIP and W\(+\) Adapter, since their reference point already includes an intermediate inversion step rather than the original source image.

\section{Conclusion}

We introduced SCOUT, a framework for discovering, validating, and manipulating concept-linked sparse features directly in
face recognition template space from natural-language queries. By combining sparse autoencoders with open-vocabulary
concept scoring, concept-agreement validation, inversion-based inspection, and feature-level intervention, SCOUT exposes
direct semantic controls without relying on an image-space edit--re-encode pipeline. Across diverse CNN- and transformer-based FR encoders, the recovered directions remain semantically coherent beyond predefined attribute taxonomies, transfer across independent inversion models, and preserve identity under intervention, comparing favorably with image-space baselines. FR templates thus admit direct, interpretable semantic control. Future work will investigate synthetic data generation from controlled template-space interventions, to produce attribute-targeted, identity-preserving variations.

\noindent\textbf{Acknowledgments.} This research was supported by the ARIS project J2-50069 (MIXBAI) and the ARIS Programme P2-0250, \emph{Metrology and Biometric Systems}.

{\small
\bibliographystyle{ieee}
\bibliography{egbib_camera_ready}
}

\end{document}